\documentclass[runningheads]{llncs}
\usepackage{graphicx}

\usepackage{tikz}
\usepackage{comment}
\usepackage{amsmath,amssymb} 
\usepackage{color}
\usepackage{bm}
\usepackage{algpseudocode}
\usepackage{algorithmicx,algorithm}
\usepackage{hyperref}
\usepackage[accsupp]{axessibility}  
\usepackage{wrapfig}
\usepackage{marvosym}

\usepackage{wrapfig}
\usepackage{subfigure}
\usepackage{caption}
\usepackage{multirow}

\begin{document}
\pagestyle{headings}
\mainmatter
\def\ECCVSubNumber{401}  

\title{Improving Vision Transformers by Revisiting High-frequency Components} 

\titlerunning{Improving Vision Transformers by Revisiting High-frequency Components}
%
%
\authorrunning{Jiawang Bai, Li Yuan, Shu-Tao Xia et al.}

\author{Jiawang Bai$^{1}$, Li Yuan$^{2,5,{\textrm{\Letter}}}$, Shu-Tao Xia$^{1,5,{\textrm{\Letter}}}$,  Shuicheng Yan$^4$, \\ Zhifeng Li$^{3,{\textrm{\Letter}}}$, and Wei Liu$^{3,{\textrm{\Letter}}}$}
\institute{$^1$Tsinghua Shenzhen International Graduate School, Tsinghua University \\$^2$School of ECE at Peking University \quad $^3$Data Platform, Tencent \\ $^4$ Sea AI Lab \quad $^5$Peng Cheng Laboratory\\
{\tt bjw19@mails.tsinghua.edu.cn; yuanli-ece@pku.edu.cn; xiast@sz.tsinghua.edu.cn; 
yansc@sea.com; \\ michaelzfli@tencent.com; wl2223@columbia.edu}}
\renewcommand{\thefootnote}{\fnsymbol{footnote}}
\footnotetext{\textrm{\Letter} Corresponding authors.}

%
\maketitle


\begin{abstract}
The transformer models have shown promising effectiveness in dealing with various vision tasks. However, compared with training Convolutional Neural Network (CNN) models, training Vision Transformer (ViT) models is more difficult and relies on the large-scale training set. To explain this observation we make a hypothesis that \textit{ViT models are less effective in capturing the high-frequency components of images than CNN models}, and verify it by a frequency analysis. Inspired by this finding, we first investigate the effects of existing techniques for improving ViT models from a new frequency perspective, and find that the success of some techniques ($e.g.$, RandAugment) can be attributed to the better usage of the high-frequency components. Then, to compensate for this insufficient ability of ViT models, we propose \texttt{HAT}, which directly augments high-frequency components of images via adversarial training. We show that \texttt{HAT} can consistently boost the performance of various ViT models ($e.g.$, +1.2\% for ViT-B, +0.5\% for Swin-B), and especially enhance the advanced model VOLO-D5 to 87.3\% that only uses ImageNet-1K data, and the superiority can also be maintained on out-of-distribution data and transferred to downstream tasks. The code is available at: \href{https://github.com/jiawangbai/HAT}{\texttt{https://github.com/jiawangbai/HAT}}.

\end{abstract}

\section{Introduction}

Recently, transformer models have shown high effectiveness in various vision tasks and attracted growing attention.
The pioneering work is Vision Transformer (ViT) \cite{dosovitskiy2020image}, which is a full-transformer architecture directly inherited from natural language processing \cite{vaswani2017attention} but taking raw image patches as input. After that, many ViT variants \cite{d2021convit,liu2021swin,chu2021twins,heo2021rethinking,wang2021pyramid,yuan2021volo,wang2022crossformer} have been proposed and achieved competitive performance with Convolutional Neural Network (CNN) models.
Though promising in vision tasks, ViT models suffer training difficulty and require significantly more training samples \cite{dosovitskiy2020image} compared with CNN models.

One reason for this difficulty may be that ViT models can not effectively exploit the local structures as they split an image to a sequence of patches and  model their dependencies with the self-attention mechanism~\cite{yuan2021tokens,park2022how}.
In contrast, CNN models can effectively extract local features within the receptive fields  \cite{brendel2018approximating,geirhos2018imagenet} with convolution operation.
From some previous studies \cite{campbell1968application,de1980spatial,sweldens1998lifting}, the local structures ($e.g.$, edges and lines) are more related to the high-frequency components of the images. We then naturally make such a hypothesis: \textit{ViT models are less effective in capturing the high-frequency components of images than CNN models.}


%




\begin{figure}[t]
\centering
\subfigure[Low-pass Filtering]{
\includegraphics[width=0.47\textwidth]{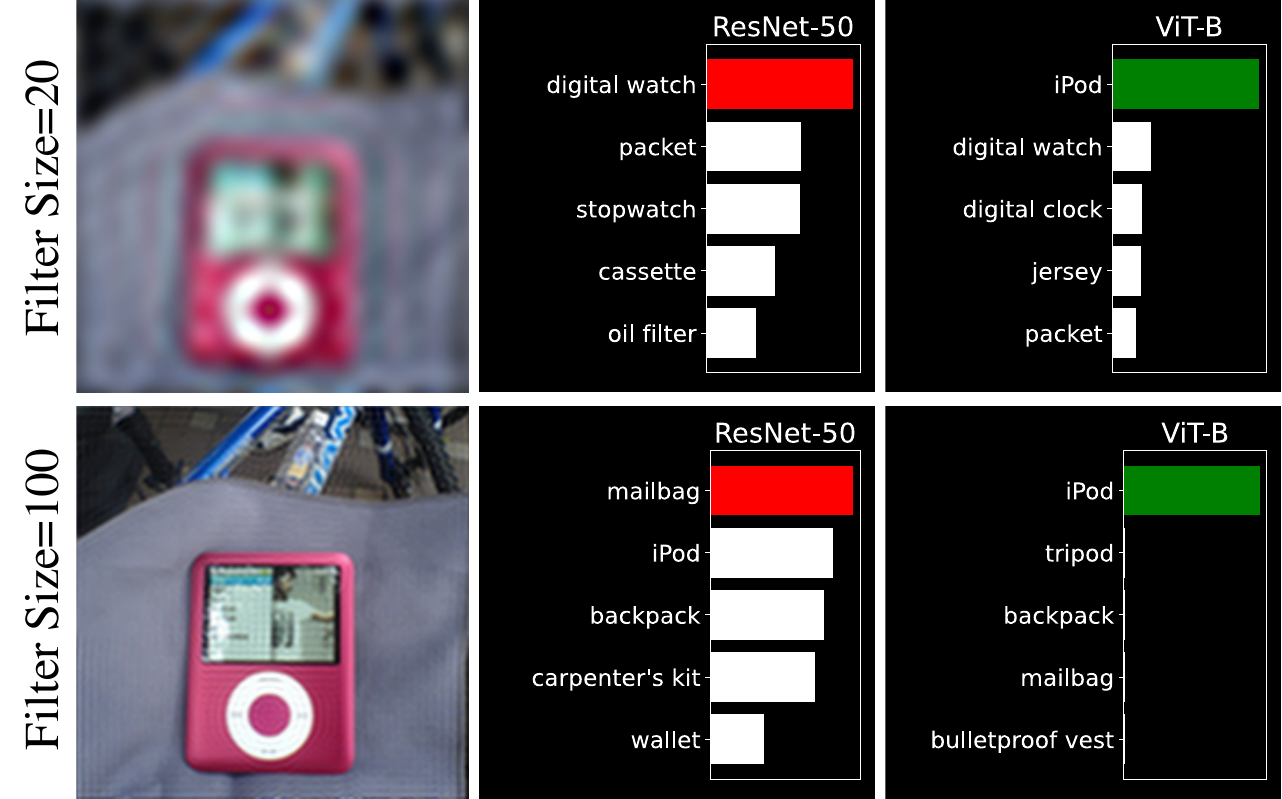}
\quad
\includegraphics[width=0.38\textwidth]{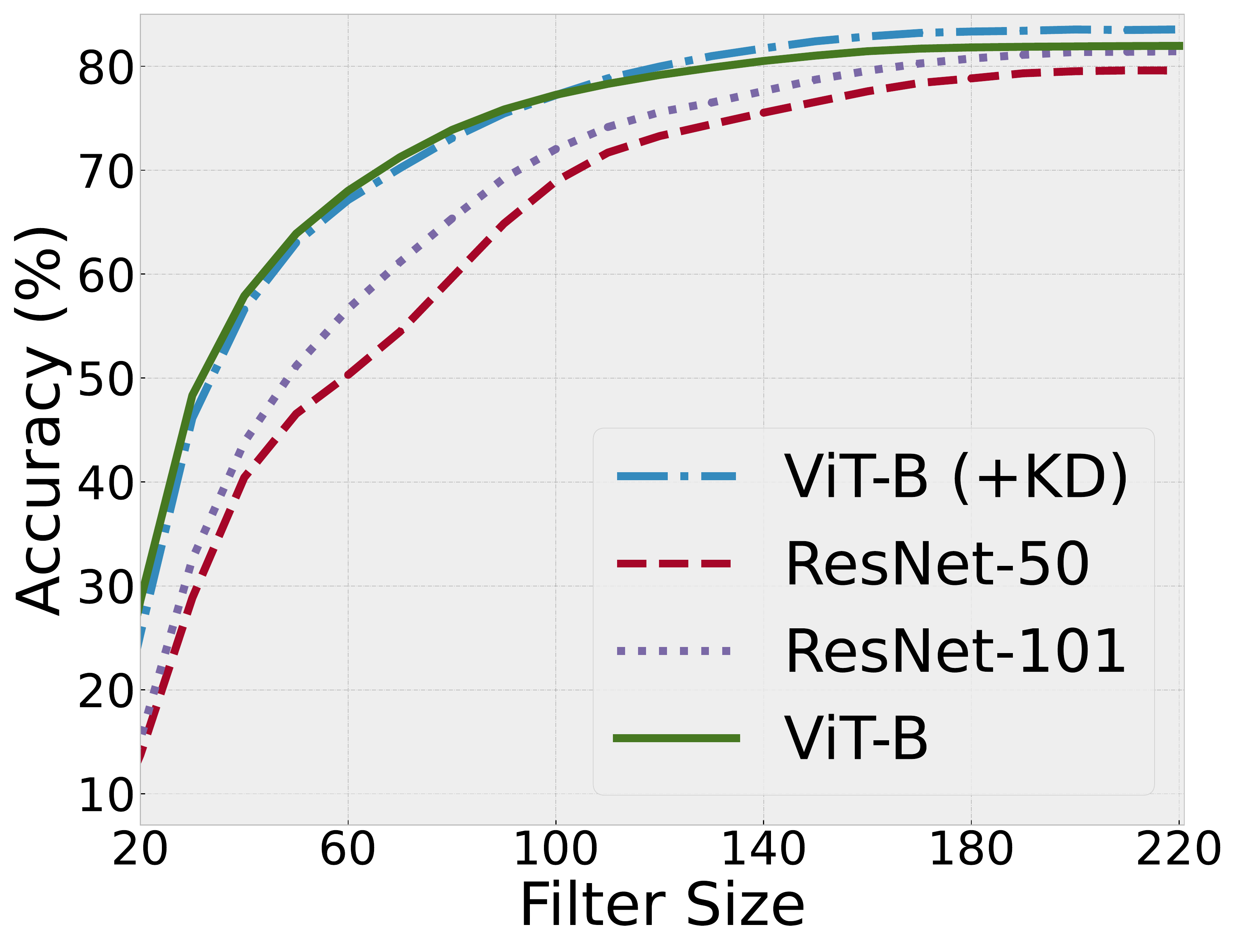}
\label{fig:vit_vs_cnn:low}
}
\subfigure[High-pass Filtering]{
\includegraphics[width=0.47\textwidth]{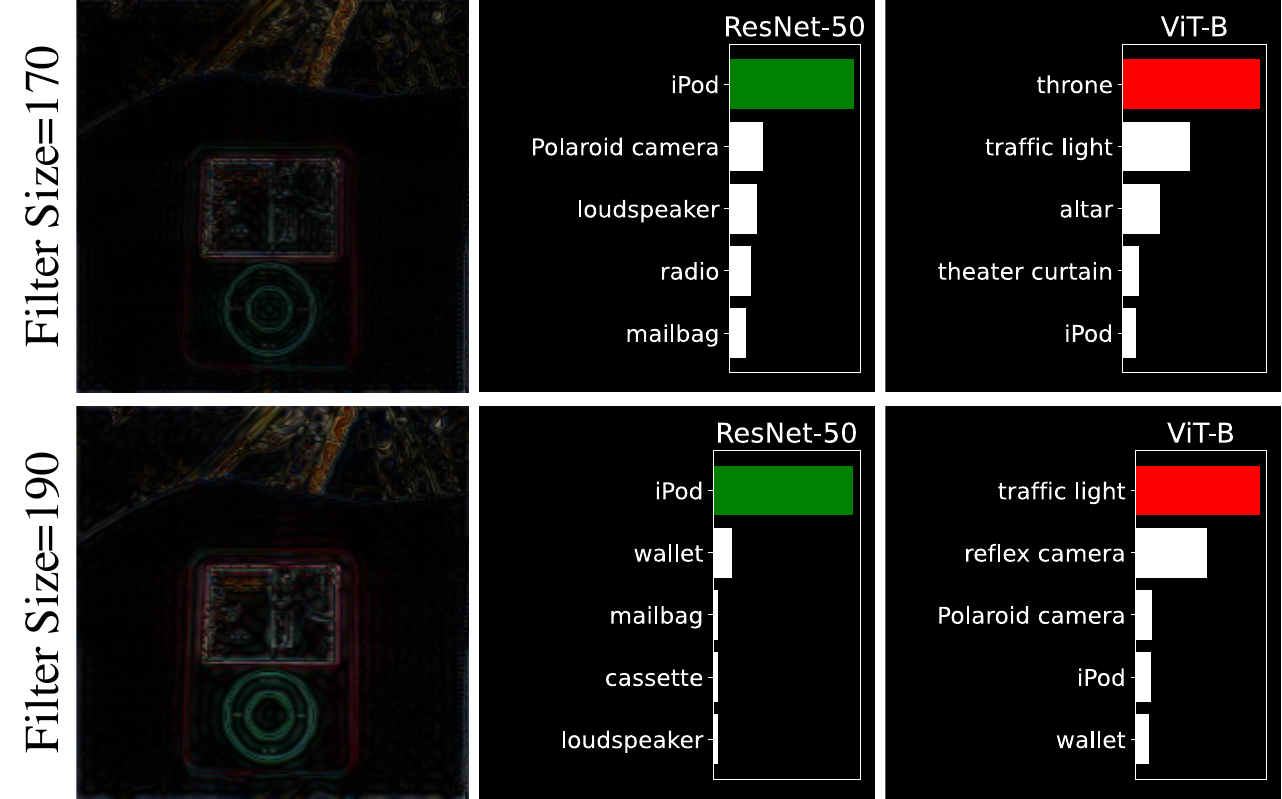}
\quad
\includegraphics[width=0.38\textwidth]{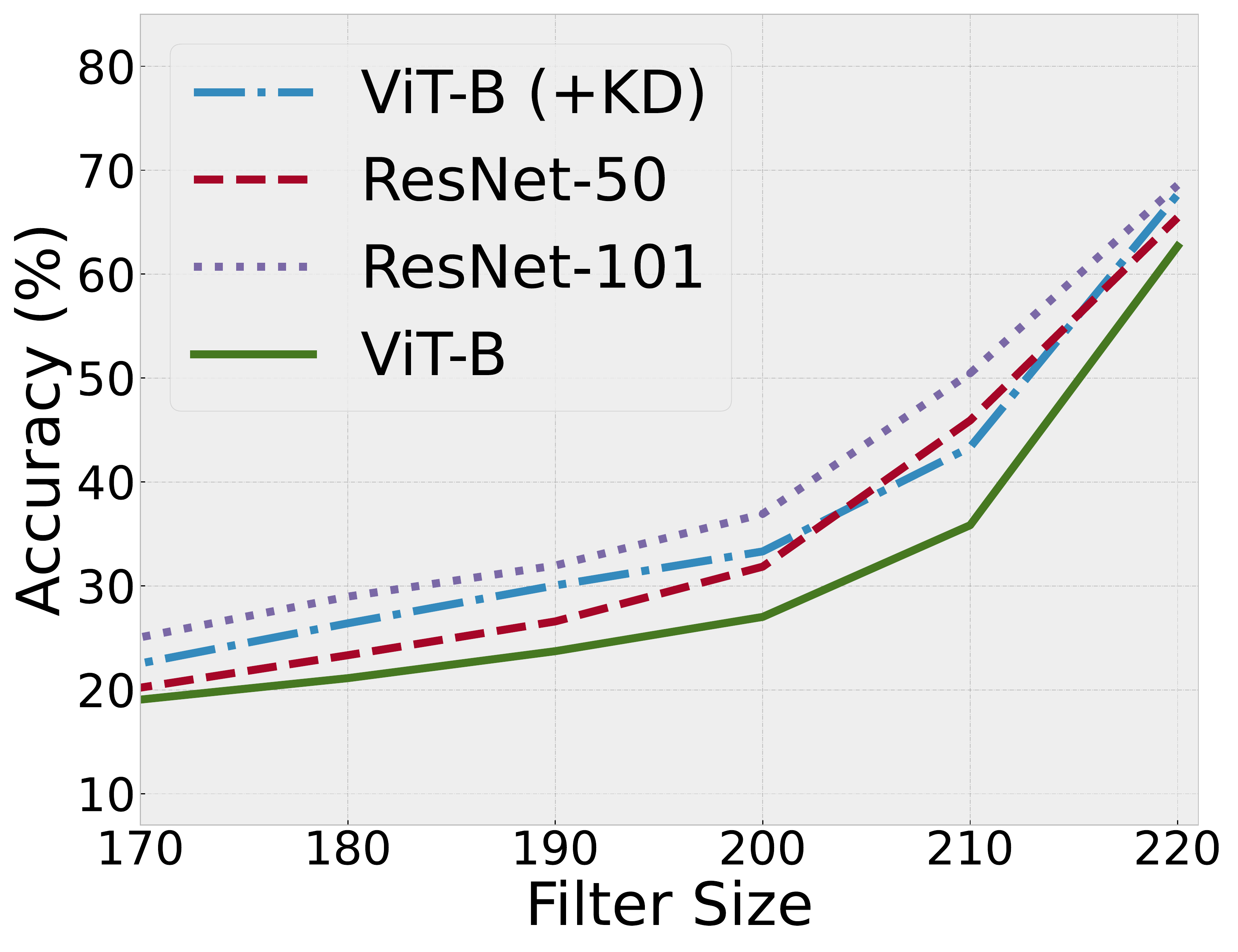}
\label{fig:vit_vs_cnn:high}
}
\setlength{\abovecaptionskip}{-0pt}
\setlength{\belowcaptionskip}{-20pt}
\caption{Comparison of ViT-B, ResNet-50, ResNet-101, and ViT-B(+KD) on low- and high-pass filtered validation set with different filter sizes. ViT-B is
the base ViT model taking as input a sequence of 16$\times$16 patches.
KD denotes knowledge distillation, where the teacher model is a RegNetY-16GF \cite{radosavovic2020designing} following \cite{touvron2021training}. The top-1 accuracy of ViT-B, ResNet-101, ResNet-50, and ViT-B (+KD) on the ImageNet validation set is 82.0\%, 79.8\%, , 81.6\%, and 83.6\%, respectively. } 
\label{fig:vit_vs_cnn}
\end{figure}

To verify our hypothesis, we use the discrete Fourier transform (DFT) to decompose the original images into the low- and high-frequency components and evaluate the model performance on them respectively \cite{yin2019fourier,wang2020high}.
Figure \ref{fig:vit_vs_cnn} shows a comparison between ViT-B \cite{dosovitskiy2020image} and ResNet-50 \cite{he2016deep}, where a larger filter size for the low- and high-pass filtering means more low- and high-frequency components, respectively.
In our experiments, ViT-B has a higher top-1 accuracy on the original ImageNet validation set (82.0\% vs. 79.8\%) and a larger model size (86.6M vs. 25.6M).
We can see that ViT-B performs better than ResNet-50 on the low-frequency components (Figure \ref{fig:vit_vs_cnn} (a)), but worse than ResNet-50 on the high-frequency components (Figure \ref{fig:vit_vs_cnn} (b)), which supports our hypothesis.

Motivated by the above observation, we study existing techniques for ViT models from a frequency perspective, including knowledge distillation \cite{jiang2021all,touvron2021training}, architecture design \cite{d2021convit,liu2021swin,chu2021twins}, and data augmentations \cite{touvron2021training,park2022how}, and provide some useful insights which are beneficial to improving performance of ViT  models.
Through extensive experiments, we find that \textit{i}) knowledge distillation is helpful to a ViT model using a CNN teacher in capturing high-frequency components of the images; \textit{ii}) compared to the original ViT \cite{dosovitskiy2020image}, some advanced architectures utilizing convolutional-like operation \cite{d2021convit} or multi-scale feature maps \cite{liu2021swin,chu2021twins} can more effectively exploit high-frequency components of the images;
\textit{iii}) RandAugment \cite{cubuk2020randaugment} is more helpful for catching high-frequency components of the images than CutMix \cite{yun2019cutmix} and Mixup \cite{zhang2018mixup}.


Furthermore, we propose to compensate for the insufficient capacity of ViT models in capturing the high-frequency components of the images by directly augmenting the high-frequency components via adversarial training \cite{madry2018towards,zhang2019theoretically}. 
Specifically, we craft adversarial examples by altering clean images with high-frequency perturbations, and jointly train ViT models over clean images and adversarial examples. Our results indicate that this training strategy improves the performance of the ViT model by compensating for its ability to capture the high-frequency components of the images. Moreover, since adversarial perturbations can naturally influence the high-frequency components in our case, we directly use adversarial perturbations without high-frequency limitation, resulting in a simple but effective method, named \texttt{HAT}, standing for improving ViT models on the \underline{h}igh-frequency components via \underline{a}dversarial \underline{t}raining. 
Note that \texttt{HAT} does not bring extra complexity during the inference stage or alter the model architecture.

Our main contributions are summarized as follows:
\begin{itemize}
  \item Based on our frequency analysis, we validate that compared to CNN models, ViT models are less effective in capturing the high-frequency components of images, which may lead to the difficulty of training ViT models.
  \item We analyze the effects of existing techniques for improving the performance of ViT models from a frequency perspective. 
  \item We propose \texttt{HAT}, which improves the performance of ViT models by influencing the high-frequency components of images directly.
  \item Our results on ImageNet classification and out-of-distribution data demonstrate the superiority of \texttt{HAT}. We also find that pre-trained models with \texttt{HAT} are beneficial to downstream tasks.
\end{itemize}

\section{Related Work}

\noindent \textbf{Transformer Models in Vision Tasks.} 
Transformer models \cite{vaswani2017attention} entirely rely on the self-attention mechanism to build long-distance dependencies, which have achieved great success in almost all natural language processing tasks \cite{DBLP:conf/naacl/DevlinCLT19,liu2019roberta,brown2020language}. Vision Transformer (ViT)  \cite{dosovitskiy2020image} is one of the earlier attempts to introduce transformer models into vision tasks, which applies a pure transformer architecture on non-overlapping image patches for image classification and has achieved  state-of-the-art accuracy. Since ViT models excel at capturing spatial information, they have also been extended to more challenging tasks, including object detection \cite{carion2020end,zhu2020deformable,dai2021up}, segmentation \cite{sun2021rethinking,strudel2021segmenter}, image enhancement \cite{chen2021pre,yang2020learning}, and video processing \cite{zhou2018end,zeng2020learning,wang2021end}. Besides, many efforts have been devoted to designing new ViT architectures \cite{d2021convit,liu2021swin,chu2021twins,heo2021rethinking,wang2021pyramid,yuan2021volo,wang2022crossformer}. For example, Liu $et$ $al$. \cite{liu2021swin} presented a hierarchical architecture with shifted window based attention that can efficiently extract multi-scale features; Yuan $et$ $al$. \cite{yuan2021volo} introduced outlook attention to efficiently encode finer-level features and contexts into tokens.

\noindent \textbf{Training Strategies for ViT Models.}
It is shown that training ViT models is more challenging than training CNN models, and requires large-scale datasets ($e.g$., ImageNet-22K \cite{deng2009imagenet} and JFT-300M \cite{sun2017revisiting}) to perform pre-training \cite{dosovitskiy2020image}.
To enable ViT to be effective on the smaller ImageNet-1K dataset \cite{deng2009imagenet}, many training strategies have been explored.
In \cite{touvron2021training,steiner2021train}, applying strong data augmentation and model regularization makes a quick solution to this problem. Among them, CutMix \cite{yun2019cutmix}, Mixup \cite{zhang2018mixup}, and RandAugment \cite{cubuk2020randaugment} are proven to be particularly helpful \cite{touvron2021training}. Besides, some customized augmentations for training ViT models are presented \cite{chen2021transmix,wang2021pyramid}. Utilizing a trained CNN teacher, knowledge distillation (KD) \cite{touvron2021training,jiang2021all} can significantly boost the performance of ViT models.
There are also some works solving this problem by using a better optimization strategy, such as promoting patch diversification \cite{gong2021vision} and sharpness-aware minimizer \cite{chen2022when}. Unlike these works, we focus on directly compensating for the ability of ViT models in capturing the high-frequency components for better performance.

\section{Revisiting ViT Models from a Frequency Perspective}
\label{sec:revisit}

To investigate ViT models from a frequency perspective, we use the discrete Fourier transform (DFT) to evaluate the model performance on certain frequency components of test samples \cite{yin2019fourier}. Let $\bm{x} \in \mathbb{R}^{H \times W}$ (omitting the dimension of image channels) and $\bm{y} \in \mathbb{R}^{C}$ represent an image in the spatial domain and its label vector, where $C$ is the number of classes. We transform $\bm{x}$ to the frequency spectrum by the DFT $\mathcal{F}: \mathbb{R}^{H \!\times\! W} \rightarrow \mathbb{C}^{H\! \times\! W}$ and transform  signals of the image from frequency back to the spatial domain by the inverse DFT $\mathcal{F}^{-1}: \mathbb{C}^{H \!\times\! W} \!\rightarrow\! \mathbb{R}^{H\! \times\! W}$. In this work, the low-frequency components are shifted to the center of the frequency spectrum. 

For a mask $\bm{m} \in \{0, 1\}^{H \times W}$, the low-pass filtering  $\mathcal{M}_l^{S}$ and high-pass filtering  $\mathcal{M}_h^{S}$ with the filter size $S$ are formally defined as:
\begin{equation}
\begin{split}
\mathcal{M}_l^{\!S}\!(\bm{x})\! =\! \mathcal{F}^{-\!1}(\bm{m}\! \odot\! \mathcal{F}(\bm{x})),
\ \text{where} \ 
\bm{m}_{i,j}\!=\!
\left\{\begin{array}{l}
1, \ \text{if} \  \min(|i\!-\!\frac{H}{2}|,|j\!-\!\frac{W}{2}|)\! \leqslant\! \frac{S}{2} \\
0, \ \text{otherwise}
\end{array}, \right.
\end{split}
\end{equation}
\begin{equation}
\begin{split}
\!\mathcal{M}_h^{\!S}\!(\bm{x})\! =\! \mathcal{F}^{-\!1}(\bm{m}\! \odot\! \mathcal{F}(\bm{x})), \text{where} \ 
\bm{m}_{i,j}\!=\!
\left\{\begin{array}{l}
0, \  \text{if} \  \min(|i\!-\!\frac{H}{2}|,|j\!-\!\frac{W}{2}|) \!\leqslant\! \frac{\min(H,W)\!-\!S}{2} \\
1, \ \text{otherwise}
\end{array}\!, \right.
\end{split}
\end{equation}
where $\odot$ is element-wise multiplication and $\bm{m}_{i,j}$ denotes the value of $\bm{m}$ at position $(i, j)$. For images containing multiple color channels, the filtering operates on each channel independently.  To make a comprehensive analysis, we evaluate various ViT architectures and training strategies with different filter sizes based on the  ImageNet validation set. We provide the visualized examples in Figure \ref{fig:vit_vs_cnn}.
 
\begin{figure}[t]
    \centering
    \subfigure[Low-pass Filtering]{
    \includegraphics[width=0.42\textwidth]{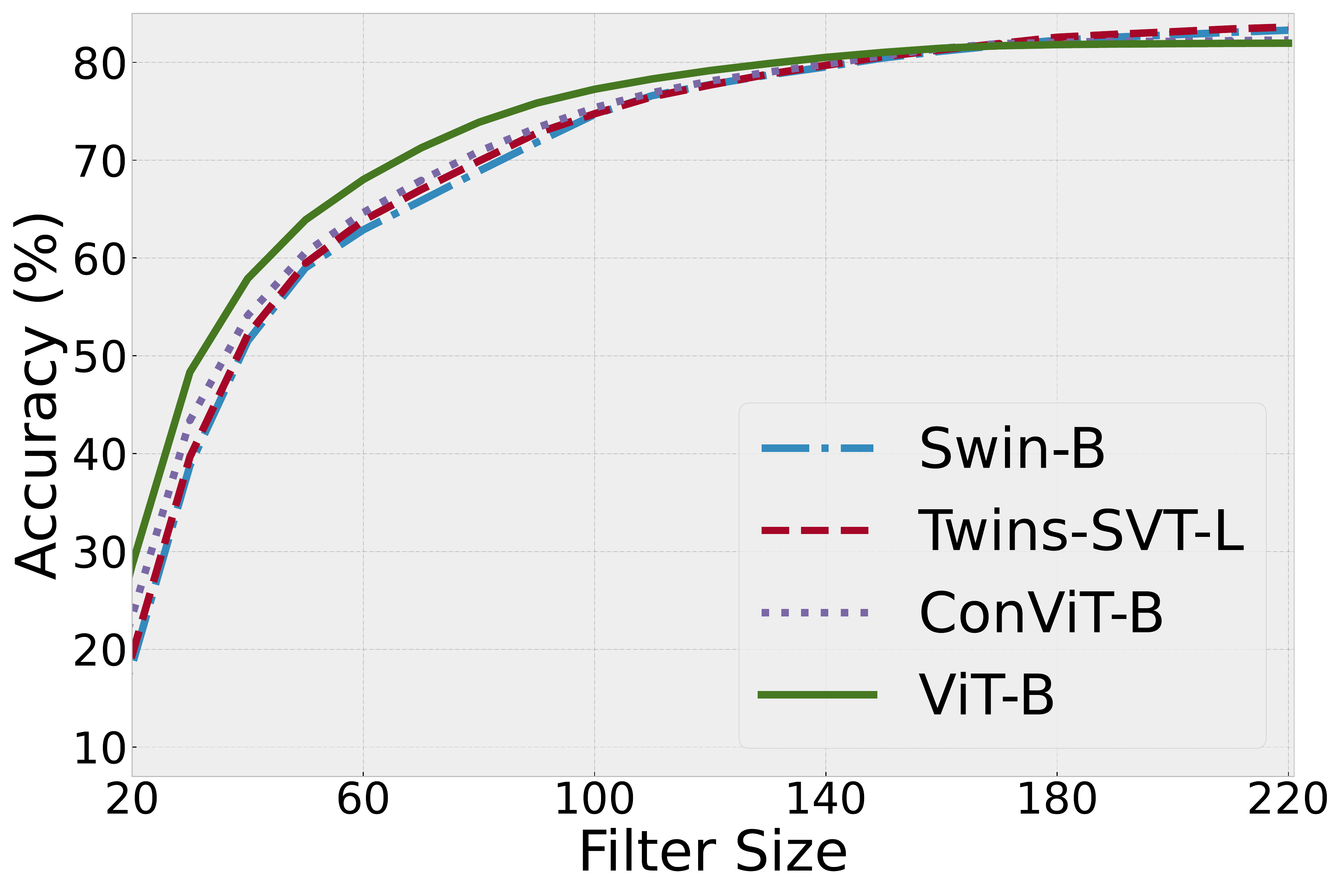}}
    \quad
    \subfigure[High-pass Filtering]{
    \includegraphics[width=0.42\textwidth]{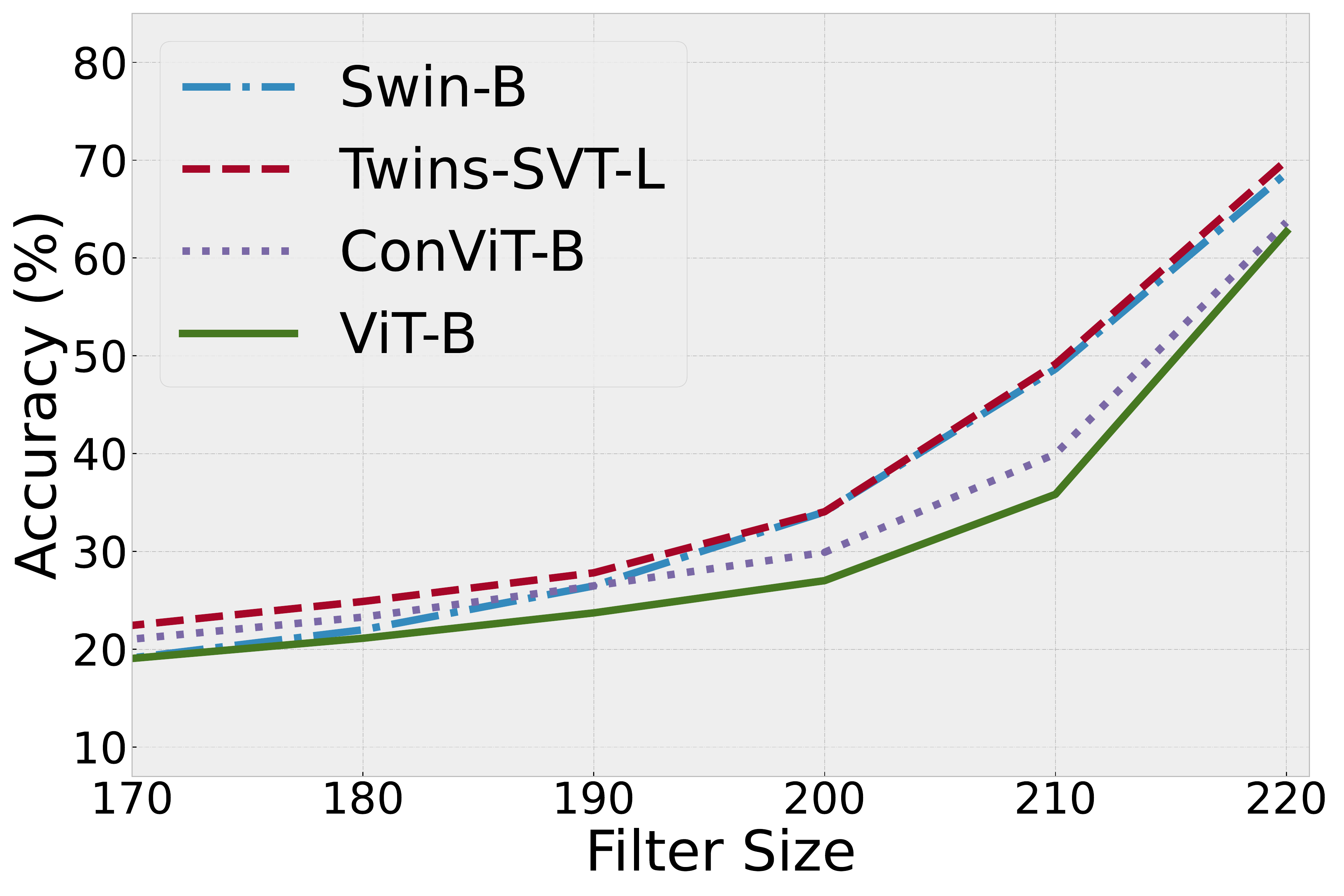}}
\setlength{\abovecaptionskip}{-0pt}
\setlength{\belowcaptionskip}{-20pt}
	\caption{Comparison of ViT-B, ConViT-B, Twins-SVT-L, and Swin-B on low- and high-pass filtered validation set with different filter sizes. The top-1 accuracy of ViT-B, ConViT-B, Twins-SVT-L, and Swin-B on the ImageNet validation set is 82.0\%, 82.3\%, 83.7\%, and 83.5\%, respectively. }
	\label{fig:vit_arch}
\end{figure}


\noindent \textbf{Comparison of ViT and CNN Models.}
Firstly, we compare the performance of ViT-B with ResNet-50 and ResNet-101, which are trained with the same data augmentations. The plots in Figure \ref{fig:vit_vs_cnn:low} show that ViT-B surpasses ResNet on the low-frequency components of images. 
However, although ViT-B achieves a higher accuracy (82.0\%) than ResNet-50 (79.8\%) and ResNet-101 (81.6\%) on the original ImageNet validation set, its performance is lower than CNN models on the high-frequency components of images, as shown in Figure \ref{fig:vit_vs_cnn:high}. 
This observation indicates that ViT models can capture the global contexts effectively, but fails to well leverage local details compared to CNN models. 
It may be because cascading self-attention blocks in ViT models is equivalent to repeatedly applying a low-pass filter, corresponding to the theoretical justification in \cite{wang2021anti}, while CNN models utilizing convolution operations behave like a series of high-pass filters \cite{park2022how} to catch more high-frequency components \cite{wang2020high}.

We further study the distillation method introduced in \cite{touvron2021training}, $i.e.$, transferring the learned knowledge in a ViT model using a CNN teacher. We use a RegNetY-16GF model \cite{radosavovic2020designing} as a teacher with the hard-label distillation, and adopt all settings in \cite{touvron2021training}. The results in Figure \ref{fig:vit_vs_cnn} show that the improvement of KD (from 82.0\% to 83.6\%) is primarily attributed to the stronger ability to exploit the high-frequency components of images. It also confirms that there is a gap between the abilities of ViT and CNN models in capturing the high-frequency components.

\noindent \textbf{Various ViT Architectures.}
Recently, various ViT architectures are proposed and show excellent results \cite{d2021convit,liu2021swin,chu2021twins}. We compare ViT-B with three advanced architectures, including ConViT-B \cite{d2021convit}, Twins-SVT-L \cite{liu2021swin}, and Swin-B \cite{chu2021twins},  with a similar model size, and present a reason for the success of these architectures from the frequency perspective. As shown in Figure \ref{fig:vit_arch}, all architectures perform similarly on the low-frequency components, while three advanced architectures achieve higher accuracy than ViT-B on the high-frequency components. Our results also provide evidence for the effects of the proposed components of these recent architectures. Specifically, the convolutional-like operation in ConViT and multi-scale feature maps in Swin Transformer and Twins-SVT can help the vision transformer capture the high-frequency components.
 
\begin{figure}[t]
    \centering
    \subfigure[Low-pass Filtering]{
    \includegraphics[width=0.42\textwidth]{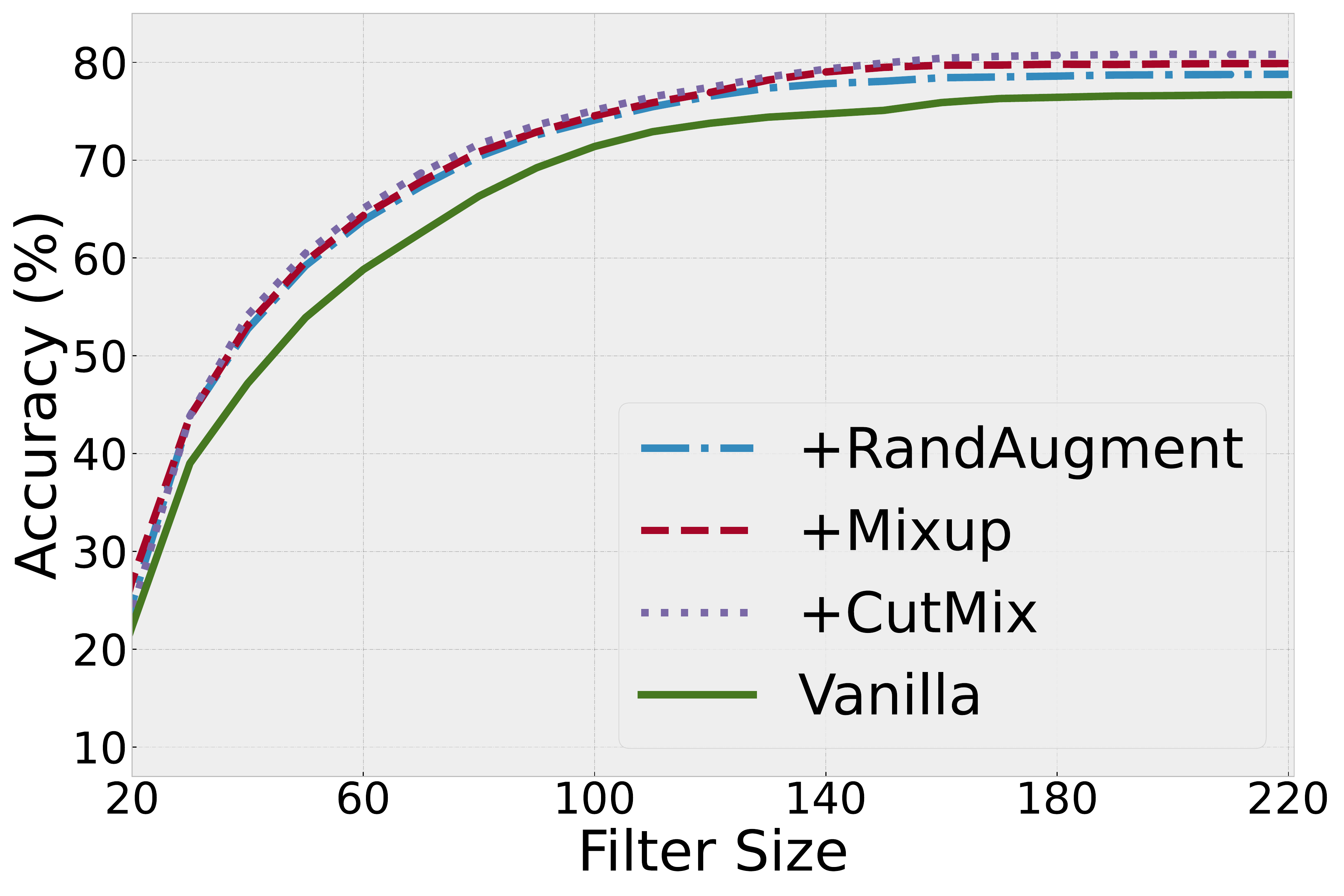}}
    \quad
    \subfigure[High-pass Filtering]{
    \includegraphics[width=0.42\textwidth]{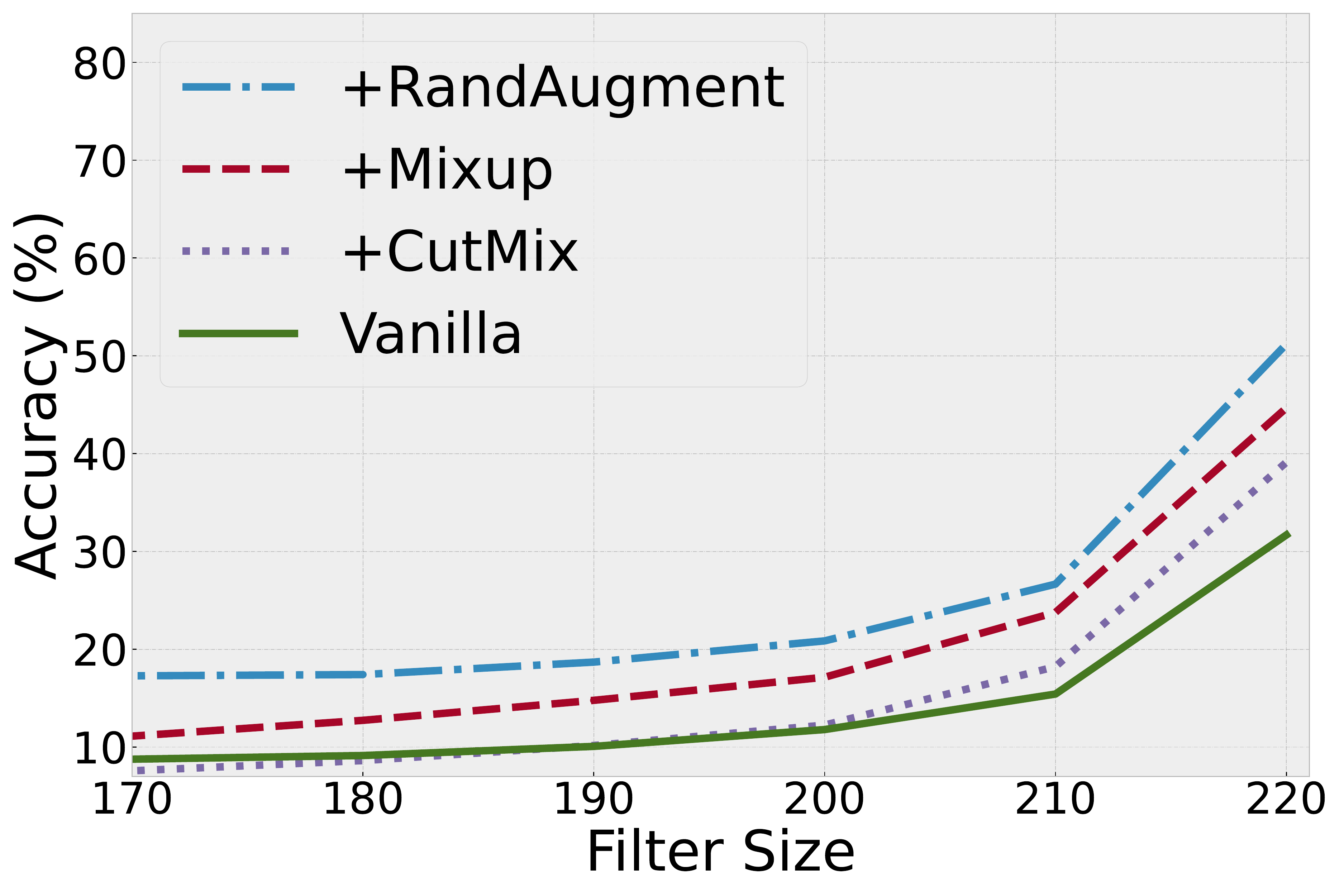}}
\setlength{\abovecaptionskip}{-0pt}
\setlength{\belowcaptionskip}{-15pt}
    \caption{Comparison of vanilla training and three data augmentations on low- and high-pass filtered validation set with different filter sizes. The top-1 accuracy of Vanilla, +CutMix, +Mixup, and +RandAugment on the ImageNet validation set is 76.7\%, 80.8\%, 79.9\%, and 78.8\%, respectively.}
    \label{fig:vit_data_aug}
\end{figure}

\noindent \textbf{Data Augmentations.}
As demonstrated in recent works \cite{touvron2021training,park2022how}, training ViT models relies heavily on strong data augmentation.  Compared to the vanilla training, the improvements of the commonly used augmentations are significant, including  CutMix \cite{yun2019cutmix} (+4.1\%), Mixup \cite{zhang2018mixup} (+3.2\%), and RandAugment \cite{cubuk2020randaugment} (+2.1\%). We make a comparison between the effects of these three augmentations from the frequency perspective. The results are shown in Figure \ref{fig:vit_data_aug}. We can see that the ranking of these three augmentations $w.r.t.$ improvements they bring is CutMix $>$ Mixup $>$ RandAugment on the low-frequency components.
However, on the high-frequency components, the case is opposite:  RandAugment $>$ Mixup $>$ CutMix. Our observation reveals that CutMix can help ViT models leverage the global context information of an image by removing a random region and replacing it with a patch from another image.  Moreover, it also indicates that the transformations used in RandAugment can force the trained model to pay more attention to high-frequency information.

\section{The Proposed Method}

In this section, we firstly describe the proposed \texttt{HAT}, and then demonstrate its effects on ViT models via a case study.

\subsection{Adversarial Training with High-frequency Perturbations}

As demonstrated by the analysis in Section \ref{sec:revisit}, the ability of ViT models to capture the high-frequency components is limited, and compensating for this limitation is a key to boosting their performance. Therefore, different from previous data augmentation methods, we propose to directly augment the high-frequency components during the training stage. 
We alter the high-frequency components of training images by adding adversarial perturbations and training ViT models on these altered images. It corresponds to adversarial training \cite{madry2018towards,zhang2019theoretically} with high-frequency perturbations and is stated formally below.

Given a ViT model $f$ with the weights $\bm{\theta}$, $f_{\bm{\theta}}(\bm{x})$ denotes its softmax output of the input sample $\bm{x}$. 
Inspired by the min-max formulation of adversarial training \cite{madry2018towards}, the objective function of adversarial training with high-frequency perturbations is as follows:
\begin{equation}
\begin{split}
\mathbb{E}_{(\bm{x},\bm{y})\! \sim \!\mathcal{D}} \!\! \left [  \! L\big(\bm{\theta}, \bm{x}, \! \bm{y}\big) \!+\! \underset{||\bm{\delta}||_\infty\! \leqslant \epsilon}{\max} \! \Big( \alpha L \big(\bm{\theta}, \bm{x}\!+\!\mathcal{M}_h^S(\bm{\delta}), \!\bm{y}\big) \!+ \!\beta L_{kl}\big(\!\bm{\theta}, \bm{x}\!+\!\mathcal{M}_h^S(\bm{\delta}),\bm{x}\big)\Big)  \!\right ],
\end{split}
\label{eq:obj_high}
\end{equation}
where $\epsilon$ denotes the maximum perturbation strength.  $L(\bm{\theta}, \bm{x}, \bm{y})\!=\!\text{CE}(f_{\bm{\theta}}(\bm{x}), \bm{y})$ and 
$L_{kl}(\bm{\theta}, \bm{x}_1, \bm{x}_2)\!=\!\frac{1}{2}[\text{KL}(f_{\bm{\theta}}(\bm{x}_1), f_{\bm{\theta}}(\bm{x}_2))\!+\!\text{KL}(f_{\bm{\theta}}(\bm{x}_2), f_{\bm{\theta}}(\bm{x}_1))]$, where CE($\cdot$) 
  \begin{wrapfigure}[14]{r}{0.47\textwidth}
    \includegraphics[width=0.47\textwidth]{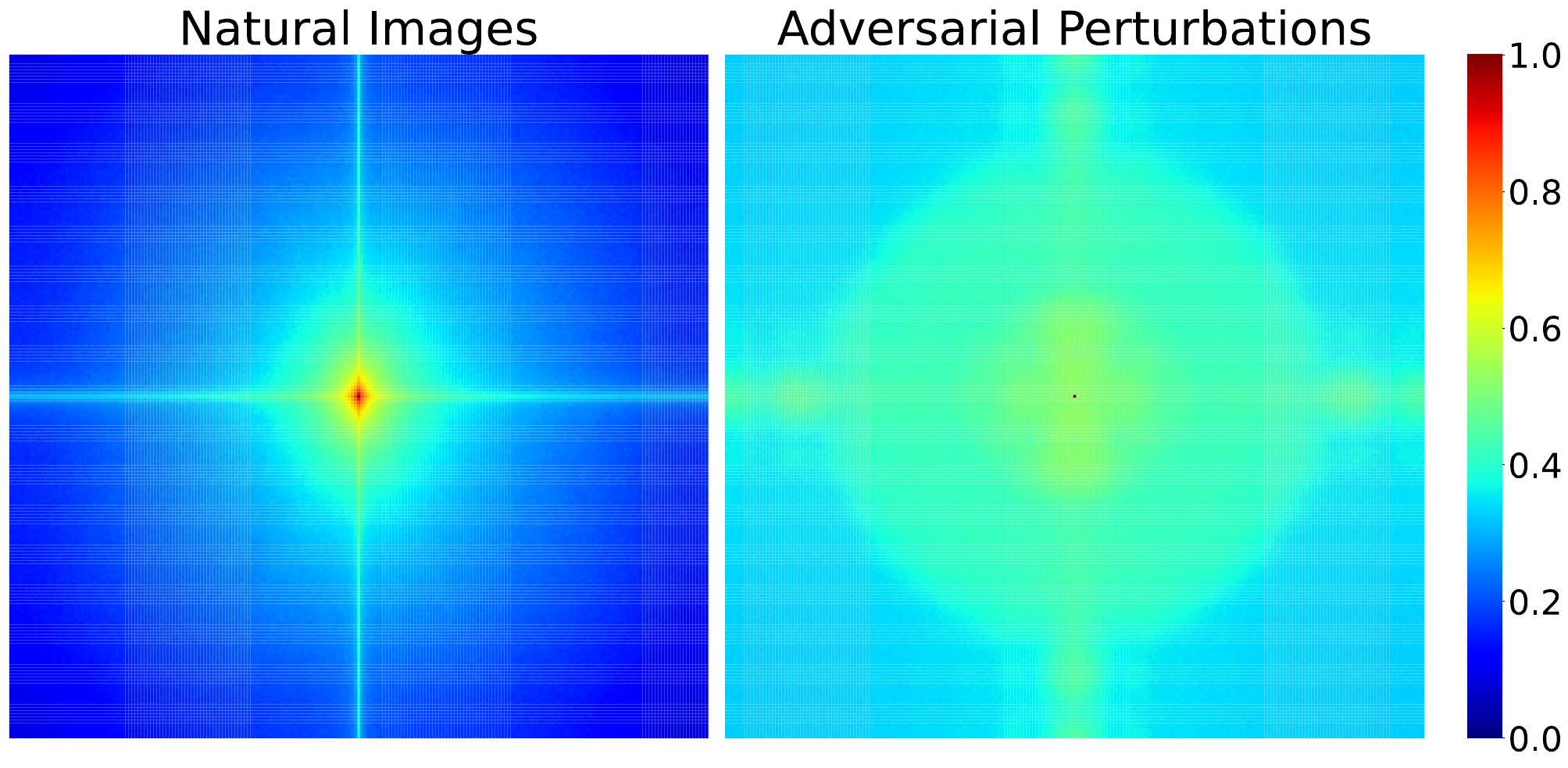}
\setlength{\abovecaptionskip}{-10pt}
    \caption{Heat maps of Fourier spectrum for natural images and adversarial perturbations. They are obtained by averaging over a batch of data. }
    \label{fig:spectrum}
\end{wrapfigure}
and KL($\cdot$) calculate the cross-entropy and the Kullback-Leibler divergence, respectively. $\alpha$ and $\beta$ are two hyper-parameters. We use the high-pass filtering  $\mathcal{M}_h^{S}$ 
with a given filter size to limit the perturbations 
in the high-frequency domain. 
Our experiments in Section \ref{sec:case_study} demonstrate that optimizing Eq. (\ref{eq:obj_high}) can compensate for 
the ability of the ViT model to capture the high-frequency components of images and thus improve its performance.

Then, we notice that adversarial perturbations are naturally imposed on the high-frequency components in our case \cite{das2018shield,liu2019feature}. It is validated in Figure \ref{fig:spectrum} that compared to natural images, adversarial perturbations show higher concentrations in the high-frequency domain. Therefore, we directly use full-frequency adversarial perturbations in our \texttt{HAT} with the below objective:
\begin{equation}
\centering
\mathbb{E}_{(\bm{x},\bm{y})\! \sim \!\mathcal{D}} \!\! \left [  L\big(\bm{\theta}, \bm{x}, \bm{y}\big) \!+\! \underset{||\bm{\delta}||_\infty\! \leqslant \epsilon}{\max} \Big( \alpha L \big(\bm{\theta}, \bm{x}\!+\!\bm{\delta}, \bm{y}\big)\! + \!\beta L_{kl}\big(\bm{\theta}, \bm{x}\!+\!\bm{\delta}, \bm{x}\big)\Big) \right ].
\label{eq:obj}
\end{equation}

The inner maximization in Eq. (\ref{eq:obj}) can be solved by project gradient descent (PGD) for $K$ steps \cite{madry2018towards}. Different from the standard PGD, following \cite{zhu2020freelb,gan2020large}, we accumulate the gradients of the model weights in each PGD step, and update the parameters at once with the accumulated gradients. In this way, the perturbations in each PGD step can be used for training. This procedure is detailed in Algorithm 1 in Appendix C. 
Besides, to address the mismatched distribution between clean images and adversarial examples \cite{xie2020adversarial}, we perform adversarial training in some initial epochs (200 epochs in our setting) and train normally in the rest epochs.

\begin{figure}[t]
\begin{minipage}{0.43\linewidth}
    \centering
    \captionof{table}{Comparison of the baseline and adversarial training (AT) with three types of perturbations, where the case of using the full-frequency perturbations corresponds to the proposed \texttt{HAT}.}
    \label{tab:at_high_low}
    \resizebox{\linewidth}{!}{
    \begin{tabular}{lc}
    \hline
    Training Strategy & Top-1 ACC (\%) \\ \hline
    Baseline & 82.0 \\
    +AT (Low-freq. Pert.) & 81.9 \\
    +AT (High-freq. Pert.) & 83.0 \\
    +AT (Full-freq. Pert.) & \textbf{83.2} \\ \hline
    \end{tabular}}
    \end{minipage}
    \quad
    \begin{minipage}{0.54\linewidth}
    \centering
    \includegraphics[width=\textwidth]{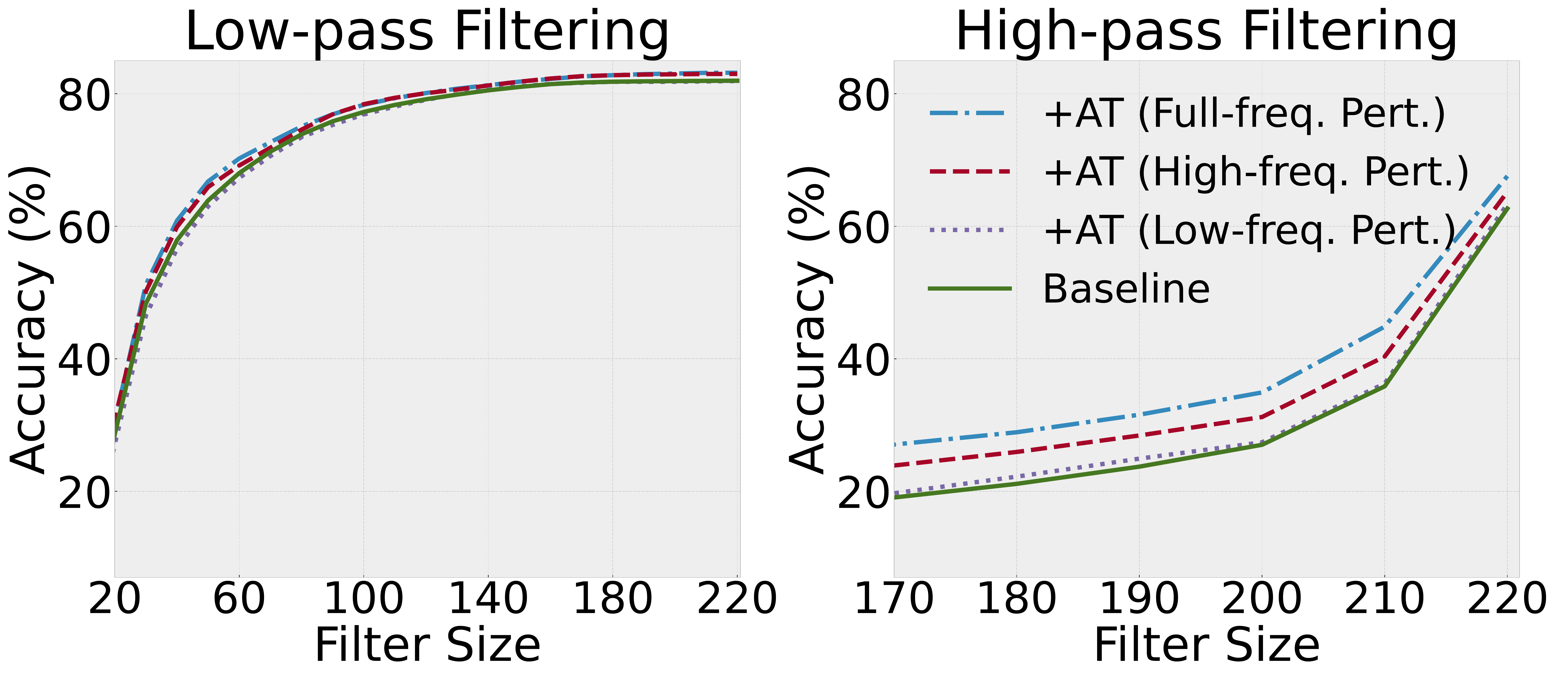}
\setlength{\abovecaptionskip}{-10pt}
	\caption{Comparison of the baseline and adversarial training (AT) with three types of perturbations on low- and high-pass filtered validation set with different filter sizes.}
    \label{fig:at_high_low}
  \end{minipage}
\end{figure}

\subsection{A Case Study using ViT-B}
\label{sec:case_study}

To illustrate how the proposed method influences the ViT models, we conduct a case study using ViT-B on ImageNet. For training ViT-B, we adopt the hyper-parameters in \cite{he2021masked} in all cases. Without considering our adversarial training, these hyper-parameters result in a strong baseline with a 82.0\% top-1 accuracy. For our method, we set $\epsilon=2/255$,  $K=3$, and $\eta=1/255$. The parameters $\alpha$ and $\beta$ are fixed at 3 and 0.01, respectively. 
We compare three types of adversarial perturbations:  low-frequency perturbations with the filter size 10, high-frequency perturbations with the filter size 10, and full-frequency perturbations.

The results on the ImageNet validation set are shown in Table \ref{tab:at_high_low}. We can see that adversarial training with the high-frequency perturbations brings 1.0\% gains over the baseline. Figure \ref{fig:at_high_low} reveals the reason for that: the ability of ViT-B to capture the high-frequency components of images is stronger than the baseline, which exactly confirms our expectation. 
In contrast, there is no improvement for the case of using low-frequency perturbations. For adversarial training with the full-frequency perturbations, without the high-filter operation and setting the filter size, it is more simple, but can also improve the performance of ViT-B. 
This case study illustrates that using the full-frequency perturbations is a reasonable choice for our \texttt{HAT}, which will be further verified in our below experiments.

\begin{table}[t]
\centering
\captionof{table}{Performance of various ViT architectures trained without and with the proposed \texttt{HAT} on the ImageNet, ImageNet-Real,
and ImageNet-V2 validation set.}
\label{tab:main_res}
\setlength\tabcolsep{4pt}
\resizebox{0.78\linewidth}{!}{
\begin{tabular}{lcccccc}
\hline
Model & Params & FLOPs & Test Size & Top-1 & \begin{tabular}[c]{@{}c@{}}Real Top-1\end{tabular} & \begin{tabular}[c]{@{}c@{}}V2 Top-1\end{tabular} \\ \hline
ViT-T & \multirow{2}{*}{5.7M} & \multirow{2}{*}{1.6G} & \multirow{2}{*}{224} & 72.2 & 80.0 & 60.1 \\
+\texttt{HAT} &  & &  & \textbf{73.3} & \textbf{81.1} & \textbf{61.0} \\ \hline
ViT-S & \multirow{2}{*}{22.1M} & \multirow{2}{*}{4.7G} & \multirow{2}{*}{224} & 80.1 & 85.7 & 68.2 \\
+\texttt{HAT} &  & &  & \textbf{80.9} & \textbf{86.6} & \textbf{70.0} \\ \hline
ViT-B & \multirow{2}{*}{86.6M} & \multirow{2}{*}{17.6G} & \multirow{2}{*}{224} & 82.0 & 87.1 & 71.0 \\
+\texttt{HAT} &  & &  & \textbf{83.2} & \textbf{87.9} & \textbf{72.6} \\ \hline
Swin-T & \multirow{2}{*}{28.3M} & \multirow{2}{*}{4.5G} & \multirow{2}{*}{224} & 81.2 & 86.8 & 70.5 \\
+\texttt{HAT} &  & &  & \textbf{82.0} & \textbf{87.3} & \textbf{71.5} \\ \hline
Swin-S & \multirow{2}{*}{49.6M} & \multirow{2}{*}{8.7G} & \multirow{2}{*}{224} & 83.0 & 87.8 & 72.4 \\
+\texttt{HAT} &  & &  & \textbf{83.3} & 87.7 & \textbf{72.8} \\ \hline
Swin-B & \multirow{2}{*}{87.8M} & \multirow{2}{*}{15.4G} & \multirow{2}{*}{224} & 83.5 & 87.9 & 72.9 \\
+\texttt{HAT} &  & &  & \textbf{84.0} & \textbf{88.2} & \textbf{73.8} \\ \hline
VOLO-D1 & \multirow{2}{*}{26.6M} & \multirow{2}{*}{6.8G} & \multirow{2}{*}{224} & 84.2 & 89.0 & 74.0 \\
+\texttt{HAT} &  & &   & \textbf{84.5} & \textbf{89.2} & \textbf{74.9} \\ \hline
VOLO-D1 & \multirow{2}{*}{26.6M} & \multirow{2}{*}{22.8G} & \multirow{2}{*}{384} & 85.2 & 89.6 & 75.6 \\
+\texttt{HAT} &  & &   & \textbf{85.5} & \textbf{89.8} & \textbf{76.6} \\ \hline
VOLO-D5 & \multirow{2}{*}{295.5M} & \multirow{2}{*}{69.0G} & \multirow{2}{*}{224} & 86.1 & 89.9 & 76.3\\
+\texttt{HAT} &  & &  & \textbf{86.3} & \textbf{90.2} & \textbf{76.8} \\ \hline
VOLO-D5 & \multirow{2}{*}{295.5M} & \multirow{2}{*}{304G} & \multirow{2}{*}{448} & 87.0 & 90.6 & 77.8 \\
+\texttt{HAT} &  & &  & \textbf{87.2} & \textbf{90.6} & \textbf{78.6} \\ \hline
VOLO-D5 & \multirow{2}{*}{295.5M} & \multirow{2}{*}{412G} & \multirow{2}{*}{512} & 87.1 & 90.6 & 78.0 \\
+\texttt{HAT} &  & &  & \textbf{87.3} & \textbf{90.7} & \textbf{78.7} \\ \hline
\end{tabular}}
\end{table}

\section{Experiments}

\subsection{Experimental Setup}
We evaluate the proposed method on ImageNet \cite{deng2009imagenet}. Our code is implemented based on PyTorch \cite{paszke2019pytorch} and  the timm library \cite{rw2019timm}. We conduct experiments on various model architectures: three variants of ViT \cite{dosovitskiy2020image} (ViT-T, ViT-S, and ViT-B with 16$\times$16 input patch size) following \cite{touvron2021training}, three variants of Swin Transformer \cite{liu2021swin} (Swin-T, Swin-S, and Swin-B), and two variants of VOLO \cite{yuan2021volo} (VOLO-D1 and VOLO-D5). ``T'', ``S'', and ``B'' denote tiny, small, and base model sizes, respectively. 
Following the standard training schedule, we train all models on the ImageNet-1K training set for 300 epochs with strong data augmentation ($e.g.$, CutMix \cite{yun2019cutmix}, Mixup \cite{zhang2018mixup}, and RandAugment \cite{cubuk2020randaugment}) and model regularization ($e.g.$, stochastic depth \cite{huang2016deep} and weight decay \cite{loshchilov2018decoupled}). 
Specifically, we use the hyper-parameters in \cite{he2021masked} for training ViT-B, and train ViT-T and ViT-S with the same hyper-parameters except for throwing away EMA, resulting in strong baselines. For training variants of Swin Transformer and VOLO, we follow the training setup of the original paper \cite{liu2021swin,yuan2021volo} (including token labeling \cite{jiang2021all} for VOLO). The default image resolution for these models is 224$\times$224. We also finetune variants of VOLO on larger image resolutions (384$\times$384, 448$\times$448, and 512$\times$512).

For the proposed \texttt{HAT},  in all cases, the PGD learning rate $\eta$ is 1/255, and the parameters $\alpha$ and $\beta$  are set to 3 and 0.01, respectively. 
We set the maximum perturbation strength $\epsilon$ as $2/255$ and the number of PGD steps $K$ as 3 by default.  For VOLO-D5, the largest model in our experiments, we set $K=2$ with $\epsilon=1/255$ to reduce the training time.  
For all ViT models, we adopt the training strategy in Algorithm 1 in the first 200 epochs and perform normal training in the rest 100 epochs. In our \texttt{HAT}, each PGD step requires one forward and backward pass. Accordingly, for the whole training, \texttt{HAT} leads to about 1.7$\times$ and 2.3$\times$  computation cost for $K=1$ and $K=2$, respectively.

\subsection{Results on ImageNet Classification}

\noindent \textbf{Results of Various ViT Architectures.} 
We present the results of variants of ViT, Swin Transformer, and VOLO trained without and with our \texttt{HAT} in Table~\ref{tab:main_res}. 
\begin{wraptable}[22]{r}{0.45\textwidth}
    \centering
\setlength{\abovecaptionskip}{-1pt}
    \captionof{table}{Comparison of \texttt{HAT} with other training strategies. All results are based on the ImageNet-1K training set and ViT-B. Results of other methods are drawn from original papers. The methods with \textcolor{blue}{blue} color mean self-supervised learning.}
    \label{tab:compare}
    \resizebox{\linewidth}{!}{
    \begin{tabular}{lccc}
    \hline
    Method & Top-1 & \begin{tabular}[c]{@{}c@{}}Real\\ Top-1\end{tabular} & \begin{tabular}[c]{@{}c@{}}V2\\ Top-1\end{tabular} \\ \hline
    Vallina & 76.7 & 82.3 & 64.1 \\
    DeiT \cite{touvron2021training} & 81.8 & 86.7 & 71.5 \\
    DeiT(+KD) \cite{touvron2021training}  & 83.4 & 88.3 & 73.2 \\
    PyramidAT \cite{herrmann2021pyramid} & 81.7 & 86.8 & 70.8 \\
    TransMix \cite{chen2021transmix} & 82.4 & - & - \\
    SAM \cite{chen2022when} & 79.9 & 85.2 & 67.5 \\
    \textcolor{blue}{DINO} \cite{caron2021emerging} & 82.8 & - & - \\
    \textcolor{blue}{MoCo v3} \cite{chen2021empirical} & 83.2 & - & - \\
    \textcolor{blue}{BEiT} \cite{bao2021beit} &  83.2 & - & - \\
    \textcolor{blue}{MAE} \cite{he2021masked} &  83.6 & - & - \\
    \texttt{HAT}(ours) & 83.2 & 87.9 & 72.6 \\
    \texttt{HAT}(+KD) & \textbf{84.3} & \textbf{88.8} & \textbf{73.9} \\ \hline
    \end{tabular}}
\end{wraptable}
``Top-1'', ``Real Top-1'', and ``V2 Top-1'' refer to the top-1 accuracy evaluated on the ImageNet~\cite{deng2009imagenet}, ImageNet-Real \cite{beyer2020we}, and ImageNet-V2 \cite{recht2019imagenet} validation set, respectively, where ImageNet-Real is built by relabeling the validation set of 
the original ImageNet for correcting labeling errors and ImageNet-V2 is a newly collected version of the ImageNet validation set.

Note that these models in Table \ref{tab:main_res} are with different architectures and sizes, and the baselines are all carefully tuned with various data augmentation and model regularization techniques. As can be seen, \texttt{HAT} can steadily improve the performance of all models.  To be specific, we boost top-1 accuracy on the ImageNet validation set by 1.1\%,  0.8\%, and 1.2\% for ViT-T, ViT-S, and ViT-B, respectively. Even for Swin Transformer and VOLO, more advanced architectures,
 we can still consistently improve the performance of their variants. The 
performance gains of our method are preserved
through finetuning at higher resolutions. In particular, when the image resolution is 512$\times$512, VOLO-D5 with our \texttt{HAT} reaches a top-1 accuracy of 87.3\% on the ImageNet validation set.

\noindent \textbf{Comparison to Other Methods with ViT-B.}
We compare our proposed \texttt{HAT} with other state-of-the-art training strategies in Table \ref{tab:compare}. We conduct these 
experiments using ViT-B. Compared to DeiT~\cite{touvron2021training} and TransMix \cite{chen2021transmix}, which utilize data augmentations to empower ViT models, we achieve a significantly higher top-1 accuracy. Most closely related to our work is pyramid adversarial training
(PyramidAT) \cite{herrmann2021pyramid}, which leverages structured adversarial perturbations. However, due to a bigger number of PGD steps, its training cost is twice as high as ours, but resulting in a lower top-1 accuracy than our \texttt{HAT}. 
Besides the supervised training methods, we also compare with the methods that pre-train on the ImageNet-1K training set in a self-supervised manner and then perform supervised finetuning, including DINO \cite{caron2021emerging}, MoCo v3 \cite{chen2021empirical}, BEiT \cite{bao2021beit}, and MAE \cite{he2021masked}. 
Our \texttt{HAT} shows very competitive performance with them. Furthermore, combining the proposed \texttt{HAT} with knowledge distillation~\cite{touvron2021training}, we obtain the performance of 84.3\%, which is the highest top-1 accuracy among these methods.

\begin{table}[t]
\centering
\caption{Performance of various ViT architectures trained without and with the proposed \texttt{HAT} on five out-of distribution datasets. Note that for mean Corruption Error (mCE), lower is better. The test resolution of all below models is 224$\times$224.}
\label{tab:ood}
\setlength\tabcolsep{2.3pt}
\begin{tabular}{lcccccccccccccc}
\hline
\multirow{2}{*}{Model} & \multicolumn{2}{c}{ImageNet-A} &  & \multicolumn{2}{c}{ImageNet-C} &  & \multicolumn{2}{c}{Sketch} &  & \multicolumn{2}{c}{Rendition} &  & \multicolumn{2}{c}{Stylized} \\ \cline{2-3} \cline{5-6} \cline{8-9} \cline{11-12} \cline{14-15} 
 & Top-1 & \begin{tabular}[c]{@{}c@{}}+\texttt{HAT}\\ Top-1\end{tabular} &  & mCE$\downarrow$ & \begin{tabular}[c]{@{}c@{}}+\texttt{HAT}\\ mCE$\downarrow$\end{tabular} &  & Top-1 & \begin{tabular}[c]{@{}c@{}}+\texttt{HAT}\\ Top-1\end{tabular} &  & Top-1 & \begin{tabular}[c]{@{}c@{}}+\texttt{HAT}\\ Top-1\end{tabular} &  & Top-1 & \begin{tabular}[c]{@{}c@{}}+\texttt{HAT}\\ Top-1\end{tabular} \\ \cline{1-3} \cline{5-6} \cline{8-9} \cline{11-12} \cline{14-15} 
ViT-T & 7.7 & 7.3 &  & 70.0 & \textbf{66.8} &  & 19.8 & \textbf{22.9} &  & 31.9 & \textbf{35.8} &  & 9.5 & \textbf{12.5} \\
ViT-S & 18.5 & \textbf{23.1} &  & 53.3 & \textbf{49.7} &  & 29.3 & \textbf{32.3} &  & 41.6 & \textbf{45.1} &  & 15.8 & \textbf{18.1} \\
ViT-B & 25.3 & \textbf{30.6} &  & 46.4 & \textbf{42.2} &  & 36.1 & \textbf{38.5} &  & 49.6 & \textbf{51.3} &  & 21.8 & \textbf{24.7} \\
Swin-T & 22.1 & \textbf{25.7} &  & 58.0 & \textbf{53.9} &  & 28.5 & \textbf{31.0} &  & 41.4 & \textbf{43.8} &  & 13.1 & \textbf{13.8} \\
Swin-S & 32.7 & \textbf{34.6} &  & 51.8 & \textbf{48.6} &  & 32.7 & \textbf{33.9} &  & 45.2 & \textbf{46.5} &  & 14.2 & \textbf{15.1} \\
Swin-B & 35.8 & \textbf{40.0} &  & 51.7 & \textbf{46.9} &  & 32.2 & \textbf{36.4} &  & 45.8 & \textbf{49.0} &  & 15.7 & \textbf{16.4} \\ 
VOLO-D1 & 39.0 & \textbf{42.9} &  & 46.8 & \textbf{43.7} &  & 38.5 & \textbf{39.5} &  & 50.3 & \textbf{51.9} &  & 19.2 & \textbf{21.2} \\ 
VOLO-D5 & 50.9 & \textbf{54.5} &  & 41.8 & \textbf{38.4} &  & 44.3 & \textbf{45.7} &  & 57.9 & \textbf{59.7} &  & 24.6 & \textbf{25.9} \\ \hline
\end{tabular}
\end{table}

\subsection{Results on Out-of-distribution Data}
We evaluate the proposed \texttt{HAT} on five out-of-distribution datasets:  
ImageNet-A which contains 7,500 examples that are harder and may cause mistakes across various models \cite{hendrycks2021natural};  
ImageNet-C \cite{hendrycks2018benchmarking} which applies a set of common visual corruptions to the ImageNet validation set; 
ImageNet-Sketch \cite{wang2019learning} which contains sketch-like images and matches the ImageNet validation set in categories and scale;  
ImageNet-Rendition \cite{hendrycks2021many}, a 30,000 image test set containing various renditions ($e.g.$, paintings, embroidery); 
Stylized ImageNet \cite{geirhos2018imagenet} which is a stylized version of ImageNet created by applying AdaIN style transfer \cite{huang2017arbitrary} to ImageNet images. We report the top-1 accuracy on all datasets, except ImageNet-C where we report the normalized mean Corruption Error (mCE) (lower is better) following the original paper \cite{hendrycks2018benchmarking}. 

The results are shown in Table \ref{tab:ood}. Note that all models are trained on the ImageNet-1K training set and tested on these five out-of-distribution datasets. As can be seen, our method can bring performance gains for all architectures, demonstrating that \texttt{HAT} can enhance the robustness of ViT models to out-of-distribution data. Accordingly, \texttt{HAT} breaks the trade-off between in-distribution and out-of-distribution generalization \cite{raghunathan2020understanding,zhang2019theoretically}, or in other words, it can achieve better performance in these two cases simultaneously.

\subsection{Transfer Learning to Downstream Tasks}

ImageNet pre-training is widely used in various vision tasks \cite{he2019rethinking}. For the downstream tasks, the backbones can be initialized by the model weights pre-trained on ImageNet. In this section, we demonstrate that the advantages of \texttt{HAT} can be transferred to the downstream tasks, including object detection, instance segmentation, and semantic segmentation. More implementation details can be found in Appendix D.


\noindent \textbf{Object Detection and Instance Segmentation.}
We take three variants of Swin Transformer trained without and with our \texttt{HAT} as pre-trained models to evaluate the performance in object detection and instance segmentation. The experiments are conducted on COCO 2017 \cite{lin2014microsoft} with the Cascade Mask R-CNN object detection framework \cite{cai2018cascade,he2017mask}. We present the results in Table~\ref{tab:det}. As can be seen, \texttt{HAT} helps three variants of Swin Transformer achieve higher detection performance. These results show that the superiority of \texttt{HAT} can be transferred to downstream tasks. Moreover, we would like to emphasize that \texttt{HAT} does not introduce extra parameters or computation cost in inference. 


\noindent \textbf{Semantic Segmentation.} We also use Swin Transformer to evaluate the performance in semantic segmentation.
We report results on the widely-used segmentation benchmark ADE20K~\cite{zhou2019semantic} with the UperNet \cite{xiao2018unified} segmentation framework.
The results are shown in Table~\ref{tab:seg}. We can see that \texttt{HAT} brings significant gains for these three variants. Especially for Swin-T and Swin-S, the improvements of \texttt{HAT} are more than 1.0\% mIOU. These results further show the benefits of our proposed \texttt{HAT} to downstream tasks. 

\begin{table}[t]
\centering
    \captionof{table}{Object detection and instance segmentation performance of the models pre-trained without and with \texttt{HAT} on COCO $val$ 2017.
    We adopt the Cascade Mask R-CNN object detection framework. $\text{AP}^{\text{box}}$ and $\text{AP}^{\text{mask}}$ are box average precision and mask average precision, respectively.}
    \label{tab:det}
    \resizebox{0.9\textwidth}{!}{
\begin{tabular}{lcccccccc}
\hline
Backbone & Params & FLOPs & $\text{AP}^{\text{box}}$ & $\text{AP}^{\text{box}}_{\text{50}}$ & $\text{AP}^{\text{box}}_{\text{75}}$ & $\text{AP}^{\text{mask}}$ & $\text{AP}^{\text{mask}}_{\text{50}}$ & $\text{AP}^{\text{mask}}_{\text{75}}$ \\ \hline
Swin-T & \multirow{2}{*}{86M} &\multirow{2}{*}{745G}& 50.5 & 69.3 & 54.9 & 43.7 & 66.6 & 47.1 \\
+\texttt{HAT} & & & \textbf{50.9} & \textbf{69.4} & \textbf{55.6} & \textbf{43.9} & \textbf{66.8} & \textbf{47.3} \\ \hline
Swin-S & \multirow{2}{*}{107M} &\multirow{2}{*}{838G}& 51.8 & 70.4 & 56.3 & 44.7 & 67.9 & 48.5 \\
+\texttt{HAT} & & & \textbf{52.5} & \textbf{71.2} & \textbf{57.1} & \textbf{45.4} & \textbf{68.8} & \textbf{49.4} \\ \hline
Swin-B & \multirow{2}{*}{145M} &\multirow{2}{*}{982G}& 51.9 & 70.9 & 56.5 & 45.0 & 68.4 & 48.7 \\
+\texttt{HAT} & & & \textbf{52.8} & \textbf{71.5} & \textbf{57.5} & \textbf{45.6} & \textbf{69.1} & \textbf{49.6} \\ \hline
\end{tabular}}
\end{table}

\begin{table}[t]
  \begin{minipage}{0.65\linewidth}
    \centering
    \captionof{table}{Semantic segmentation performance of the models pre-trained without and with \texttt{HAT} on the ADE20K validation set. We adopt the UperNet segmentation framework. MS denotes testing with variable input size.}
    \label{tab:seg}
    \resizebox{0.9\linewidth}{!}{
    \begin{tabular}{lccccc}
\hline
Backbone & Params & FLOPs & mIoU & mIoU(MS) & mAcc \\ \hline
Swin-T & \multirow{2}{*}{60M} &\multirow{2}{*}{945G} &44.5 & 46.1 & 55.6 \\
+\texttt{HAT} & & & \textbf{45.6} & \textbf{46.7} & \textbf{57.4} \\ \hline
Swin-S & \multirow{2}{*}{81M} &\multirow{2}{*}{1038G}& 47.6 & 49.5 & 58.8 \\
+\texttt{HAT}  & & & \textbf{48.1} & \textbf{49.7} & \textbf{59.5} \\ \hline
Swin-B & \multirow{2}{*}{121G} &\multirow{2}{*}{1088G}& 48.1 & 49.7 & 59.1 \\
+\texttt{HAT}  & & & \textbf{48.9} & \textbf{50.3} & \textbf{60.2} \\ \hline
\end{tabular}}
  \end{minipage}
  \hfill
  \begin{minipage}{0.32\linewidth}
    \centering
    \captionof{table}{Performance of ViT-B trained with the proposed \texttt{HAT} under different maximum perturbation strength $\epsilon$.}
    \label{tab:eps}
    \resizebox{\linewidth}{!}{
\begin{tabular}{cccc}
\hline
$ \ \ \epsilon$ \ & Top-1 & \begin{tabular}[c]{@{}c@{}}Real\\ Top-1\end{tabular} & \begin{tabular}[c]{@{}c@{}}V2\\ Top-1\end{tabular} \\ \hline
 1/255  & 83.1 & 87.8 & 72.4 \\
 2/255  & \textbf{83.2} & 87.9 & \textbf{72.6} \\
 3/255  & 83.1 & \textbf{88.0} & 72.5 \\
 4/255  & 83.1 & 87.9 & 72.5 \\
 5/255  & 83.0 & 87.9 & 72.3 \\ \hline
\end{tabular}}
  \end{minipage}
\end{table}

\subsection{Ablation Studies}
\noindent \textbf{Ablation on Maximum Perturbation Strength.}
We ablate the effects of the maximum perturbation strength $\epsilon$, where a larger $\epsilon$ indicates stronger adversarial examples for the adversarial training. We test \texttt{HAT} with $\epsilon \in$ $\{1/255, 2/255,$ $3/255,$ $4/255,$ $5/255\}$. Correspondingly, we set the number of PGD steps as $K \in \{2, 3, 4, 5, 6\}$. The PGD learning rate $\eta$ is fixed at 1/255. The results are shown in Table \ref{tab:eps}. We can see that \texttt{HAT} can achieve superior performance when $\epsilon$ is set as 2/255 or 3/255. It demonstrates that medium strength adversarial examples are more helpful for training ViT models than weaker or stronger adversarial examples. This ablation study and other experiments in this paper verify that $\epsilon=2/255$ and $K=3$ are reasonable choices in most cases.

\begin{figure}[t]

  \begin{minipage}{0.32\linewidth}
\centering
\includegraphics[width=\linewidth]{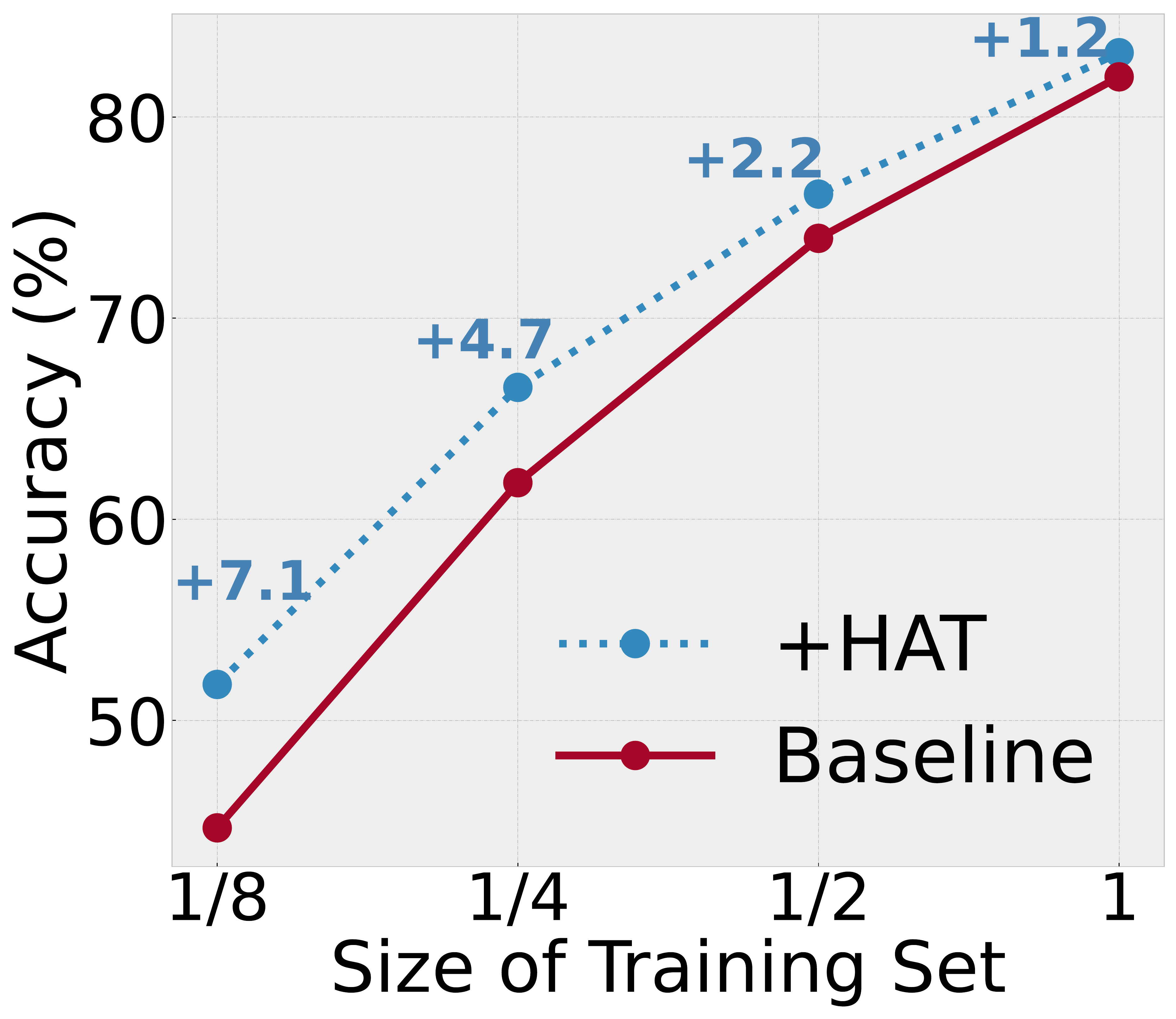}
\setlength{\abovecaptionskip}{-7pt}
    \caption{Performance of ViT-B trained without and with the proposed \texttt{HAT} on various sizes of training sets.}
    \label{fig:data_scale}
  \end{minipage}
  \hfill
  \begin{minipage}{0.65\linewidth}
    \centering
    \includegraphics[width=0.85\linewidth]{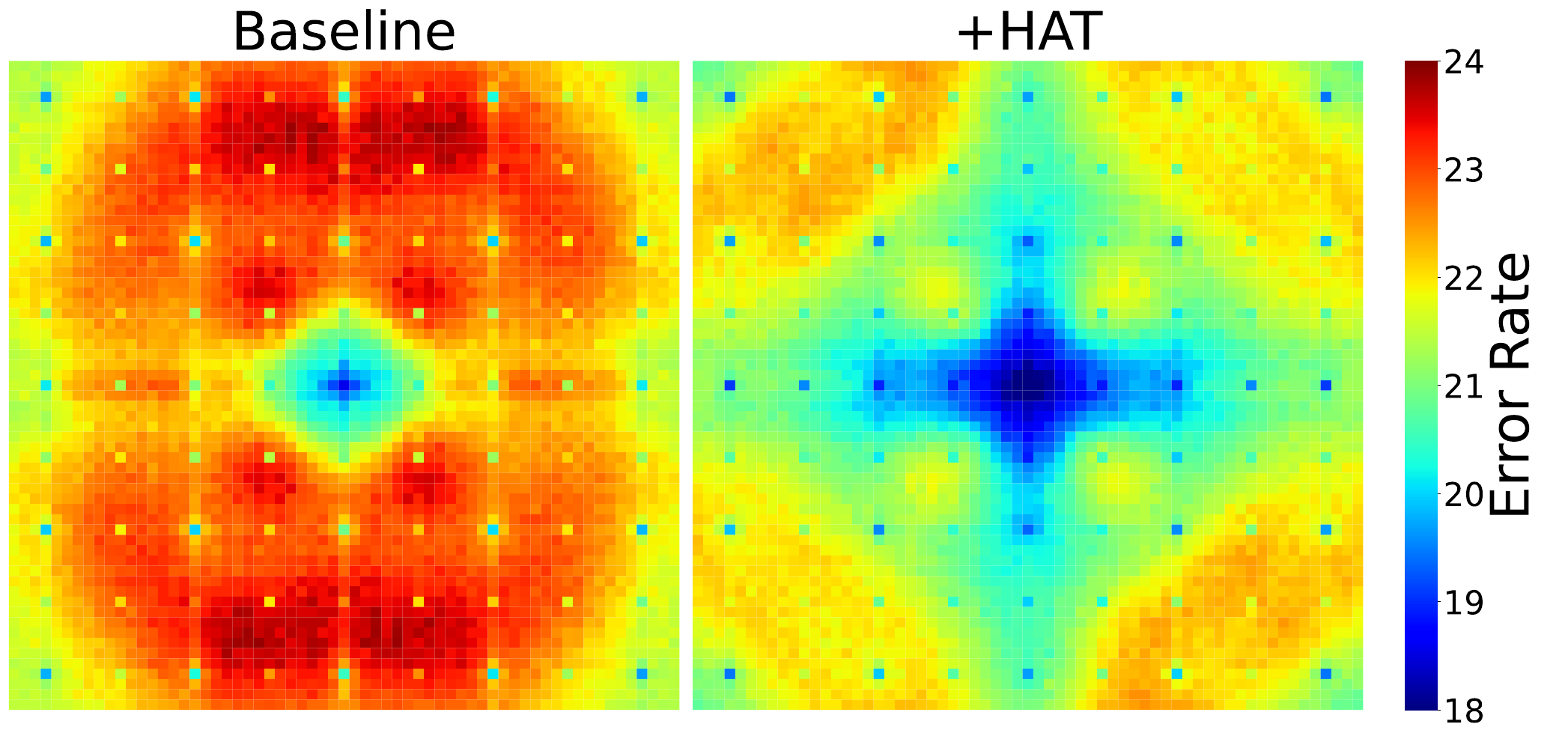}
    \caption{Fourier heat maps of ViT-B trained without and with \texttt{HAT}. The Fourier heat map reflects the sensitivity of a model to high- and low-frequency corruptions. Error rates are averaged over the entire ImageNet validation set.}
    \label{fig:model_heatmap}
  \end{minipage}
\end{figure}

\noindent \textbf{Scaling the Training Set Size.}
\texttt{HAT} 
In this part, we investigate the effects of 
with various training set sizes. Specifically, we experiment on training sets with 
different sizes by randomly sampling 1/8, 1/4, and 1/2 images from the original ImageNet-1K training set, resulting in training sets with 16K, 32K, and 64K samples. Then, we train ViT-B on these datasets without and with  the proposed \texttt{HAT} and evaluate on the original ImageNet validation set. The comparison is presented in Figure~\ref{fig:data_scale}. It is shown that \texttt{HAT} can improve the performance of ViT-B under all sizes of the training set. Especially for the small-scale training set, \texttt{HAT} brings more gains, $e.g.$, +7.1 for the smallest training set. It may be because the insufficient ability of ViT models to capture high-frequency components is amplified on the small-scale dataset and the effectiveness of \texttt{HAT} on compensating for this ability is more significant in this case. In short, our results illustrate that \texttt{HAT} enables ViT models to handle the small-scale training set better.


\subsection{Discussions}

\noindent \textbf{Fourier Heat Maps of ViT Models.}
We investigate the effects of \texttt{HAT} on the model sensitivity to low- and high-frequency corruptions. We adopt the Fourier heat map \cite{yin2019fourier}, which visualizes the error rates of a model tested on perturbed images with additive Fourier basis noise. We fix $\ell_2$-norm of the additive noise as 15.7 and average the error rates over the entire ImageNet validation set. Following \cite{yin2019fourier}, we present the 63$\times$63 square centered at the lowest frequency in the Fourier domain. The Fourier heat maps of ViT-B trained without and with \texttt{HAT} are shown in Figure \ref{fig:model_heatmap}. As we can see, the baseline model is highly sensitive to additive noise in the high-frequency. In contrast, the model trained with \texttt{HAT} is more robust to the noise, especially in the high-frequency.

\begin{table}[t]
  \begin{minipage}{0.46\linewidth}
    \centering
    \caption{\footnotesize{Comparison of \texttt{HAT} and longer normal training with ViT-B under
various computational costs.}}
\label{tab:lnt}
\centering
\setlength{\tabcolsep}{1mm}
    \resizebox{\linewidth}{!}{
\begin{tabular}{lccccc}
\hline
\multirow{2}{*}{Cost} & \multicolumn{2}{c}{Normal Training} &  & \multicolumn{2}{c}{+\texttt{HAT}} \\ \cline{2-3} \cline{5-6} 
 & Setting & Top-1 &  & Setting & Top-1 \\ \hline
1$\times$ &  300 epochs & 82.0 &  & - & - \\
1.7$\times$ & 500 epochs & 82.5 &  & $K\!=\!2$ & \textbf{83.1} \\
2.3$\times$ & 690 epochs & 82.4 &  & $K\!=\!3$ & \textbf{83.2} \\ \hline
\end{tabular}}
  \end{minipage}
  \hfill
  \begin{minipage}{0.53\linewidth}
    \centering
    \caption{Performance of CNN (ResNet-50) and MLP  (ViP-Small/7) models trained without and with the proposed \texttt{HAT}.}
    \label{tab:vip_resnet}
\setlength{\tabcolsep}{1mm}
    \resizebox{0.9\linewidth}{!}{
\begin{tabular}{lccc}
\hline
Model & Top-1 & \begin{tabular}[c]{@{}c@{}}Real Top-1\end{tabular} & \begin{tabular}[c]{@{}c@{}}V2 Top-1\end{tabular} \\ \hline
ResNet-50 & 79.8 & 86.9 & 70.7 \\
+\texttt{HAT} & \textbf{80.2} & \textbf{87.2} & \textbf{71.3} \\ \hline
ViP-Small/7 & 81.6 & 85.5 & 67.8 \\
+\texttt{HAT} & \textbf{82.2} & \textbf{86.1} & \textbf{69.6} \\ \hline
\end{tabular}}
  \end{minipage}
\end{table}

\noindent \textbf{Longer Normal Training.} To further verify the effectiveness of \texttt{HAT}, we increase the number of epochs of normal training to match the computational cost of \text{HAT}. The results are presented in Table \ref{tab:lnt}. We can see that 500 epochs are enough for the normal training to converge, and \texttt{HAT} surpasses normal training under 1.7$\times$ and 2.3$\times$ cost.

\noindent \textbf{Beyond ViT Models.}
We explore the performance of \texttt{HAT} on the CNN and MLP models. We conduct experiments on ResNet-50 \cite{he2016deep} and ViP-Small-7~\cite{hou2022vision}. We train ResNet-50 for 800 epochs following the setup in \cite{wightman2021resnet} and  ViP-Small-7 for 300 epochs following the setup in \cite{hou2022vision}. For our \texttt{HAT}, we perform adversarial training in the first 600 epochs for ResNet-50 and in the first 200 epochs for ViP-Small-7, and keep other settings unchanged. The results in Table~\ref{tab:vip_resnet} show that \texttt{HAT}  brings improvements of 0.4\% and 0.6\% for ResNet-50 and ViP-Small-7, respectively. Therefore, the proposed training strategy is promising to be extended to other models.



\section{Conclusions and Future Work}

In this paper, we study ViT models from a frequency perspective. We find that compared to CNN models, ViT models can not well exploit the high-frequency components of images. We also present a new frequency analysis of existing techniques for improving the performance of ViT models. To compensate for this insufficient ability of ViT models, we propose \texttt{HAT}, a simple but effective training strategy based on adversarial training. Extensive experiments verify its effectiveness on diverse benchmarks.

Despite achieving higher performance, \texttt{HAT} has 
an increased training time compared to the normal training. Therefore, a future study is to improve the efficiency of the proposed  \texttt{HAT}. Also, the insights provided in this paper further prompt us to explore other techniques to compensate for the ability of ViT models to capture the high-frequency components of images.

\noindent \textbf{Acknowledgments.}
This work is supported in part by the National Natural Science Foundation of China under Grant 62171248, and the PCNL KEY project (PCL2021A07).
Li Yuan was supported in part by PKU-Shenzhen Start-Up  Research Fund (1270110283) and PengCheng Laboratory.

\clearpage
%
%
\bibliographystyle{splncs04}
\bibliography{egbib}

\begin{thebibliography}{10}
\providecommand{\url}[1]{\texttt{#1}}
\providecommand{\urlprefix}{URL }
\providecommand{\doi}[1]{https://doi.org/#1}

\bibitem{alquraishi2019alphafold}
AlQuraishi, M.: Alphafold at casp13. Bioinformatics  \textbf{35}(22),
  4862--4865 (2019)

\bibitem{bao2021beit}
Bao, H., Dong, L., Wei, F.: Beit: Bert pre-training of image transformers.
  arXiv preprint arXiv:2106.08254  (2021)

\bibitem{beyer2020we}
Beyer, L., H{\'e}naff, O.J., Kolesnikov, A., Zhai, X., Oord, A.v.d.: Are we
  done with imagenet? arXiv preprint arXiv:2006.07159  (2020)

\bibitem{brendel2018approximating}
Brendel, W., Bethge, M.: Approximating cnns with bag-of-local-features models
  works surprisingly well on imagenet. In: ICLR (2019)

\bibitem{brown2020language}
Brown, T., Mann, B., Ryder, N., Subbiah, M., Kaplan, J.D., Dhariwal, P.,
  Neelakantan, A., Shyam, P., Sastry, G., Askell, A., et~al.: Language models
  are few-shot learners. In: NeurIPS (2020)

\bibitem{cai2018cascade}
Cai, Z., Vasconcelos, N.: Cascade r-cnn: Delving into high quality object
  detection. In: CVPR (2018)

\bibitem{campbell1968application}
Campbell, F.W., Robson, J.G.: Application of fourier analysis to the visibility
  of gratings. The Journal of physiology  \textbf{197}(3), ~551 (1968)

\bibitem{carion2020end}
Carion, N., Massa, F., Synnaeve, G., Usunier, N., Kirillov, A., Zagoruyko, S.:
  End-to-end object detection with transformers. In: ECCV (2020)

\bibitem{caron2021emerging}
Caron, M., Touvron, H., Misra, I., J{\'e}gou, H., Mairal, J., Bojanowski, P.,
  Joulin, A.: Emerging properties in self-supervised vision transformers. In:
  ICCV (2021)

\bibitem{chen2021pre}
Chen, H., Wang, Y., Guo, T., Xu, C., Deng, Y., Liu, Z., Ma, S., Xu, C., Xu, C.,
  Gao, W.: Pre-trained image processing transformer. In: CVPR (2021)

\bibitem{chen2021transmix}
Chen, J.N., Sun, S., He, J., Torr, P., Yuille, A., Bai, S.: Transmix: Attend to
  mix for vision transformers. arXiv preprint arXiv:2111.09833  (2021)

\bibitem{chen2022when}
Chen, X., Hsieh, C.J., Gong, B.: When vision transformers outperform resnets
  without pre-training or strong data augmentations. In: ICLR (2022)

\bibitem{chen2021empirical}
Chen, X., Xie, S., He, K.: An empirical study of training self-supervised
  vision transformers. In: ICCV (2021)

\bibitem{chu2021twins}
Chu, X., Tian, Z., Wang, Y., Zhang, B., Ren, H., Wei, X., Xia, H., Shen, C.:
  Twins: Revisiting the design of spatial attention in vision transformers. In:
  NeurIPS (2021)

\bibitem{cubuk2020randaugment}
Cubuk, E.D., Zoph, B., Shlens, J., Le, Q.V.: Randaugment: Practical automated
  data augmentation with a reduced search space. In: CVPR Workshops (2020)

\bibitem{dai2021up}
Dai, Z., Cai, B., Lin, Y., Chen, J.: Up-detr: Unsupervised pre-training for
  object detection with transformers. In: CVPR (2021)

\bibitem{das2018shield}
Das, N., Shanbhogue, M., Chen, S.T., Hohman, F., Li, S., Chen, L., Kounavis,
  M.E., Chau, D.H.: Shield: Fast, practical defense and vaccination for deep
  learning using jpeg compression. In: ACM SIGKDD (2018)

\bibitem{de1980spatial}
De~Valois, R.L., De~Valois, K.K.: Spatial vision. Annual review of psychology
  \textbf{31}(1),  309--341 (1980)

\bibitem{deng2009imagenet}
Deng, J., Dong, W., Socher, R., Li, L.J., Li, K., Fei-Fei, L.: Imagenet: A
  large-scale hierarchical image database. In: CVPR (2009)

\bibitem{deng2019mutual}
Deng, Z., Peng, X., Li, Z., Qiao, Y.: Mutual component convolutional neural
  networks for heterogeneous face recognition. IEEE Transactions on Image
  Processing  \textbf{28}(6),  3102--3114 (2019)

\bibitem{DBLP:conf/naacl/DevlinCLT19}
Devlin, J., Chang, M.W., Lee, K., Toutanova, K.: Bert: Pre-training of deep
  bidirectional transformers for language understanding. In: NAACL (2019)

\bibitem{devlin2019bert}
Devlin, J., Chang, M.W., Lee, K., Toutanova, K.: Bert: Pre-training of deep
  bidirectional transformers for language understanding. In: NAACL-HLT (1)
  (2019)

\bibitem{dosovitskiy2020image}
Dosovitskiy, A., Beyer, L., Kolesnikov, A., Weissenborn, D., Zhai, X.,
  Unterthiner, T., Dehghani, M., Minderer, M., Heigold, G., Gelly, S., et~al.:
  An image is worth 16x16 words: Transformers for image recognition at scale.
  In: ICLR (2020)

\bibitem{d2021convit}
d’Ascoli, S., Touvron, H., Leavitt, M.L., Morcos, A.S., Biroli, G., Sagun,
  L.: Convit: Improving vision transformers with soft convolutional inductive
  biases. In: ICML (2021)

\bibitem{gan2020large}
Gan, Z., Chen, Y.C., Li, L., Zhu, C., Cheng, Y., Liu, J.: Large-scale
  adversarial training for vision-and-language representation learning. In:
  NeurIPS (2020)

\bibitem{geirhos2018imagenet}
Geirhos, R., Rubisch, P., Michaelis, C., Bethge, M., Wichmann, F.A., Brendel,
  W.: Imagenet-trained cnns are biased towards texture; increasing shape bias
  improves accuracy and robustness. In: ICLR (2018)

\bibitem{gong2021vision}
Gong, C., Wang, D., Li, M., Chandra, V., Liu, Q.: Vision transformers with
  patch diversification. arXiv preprint arXiv:2104.12753  (2021)

\bibitem{he2021masked}
He, K., Chen, X., Xie, S., Li, Y., Doll{\'a}r, P., Girshick, R.: Masked
  autoencoders are scalable vision learners. arXiv preprint arXiv:2111.06377
  (2021)

\bibitem{he2019rethinking}
He, K., Girshick, R., Doll{\'a}r, P.: Rethinking imagenet pre-training. In:
  CVPR (2019)

\bibitem{he2017mask}
He, K., Gkioxari, G., Doll{\'a}r, P., Girshick, R.: Mask r-cnn. In: ICCV (2017)

\bibitem{he2016deep}
He, K., Zhang, X., Ren, S., Sun, J.: Deep residual learning for image
  recognition. In: CVPR (2016)

\bibitem{hendrycks2021many}
Hendrycks, D., Basart, S., Mu, N., Kadavath, S., Wang, F., Dorundo, E., Desai,
  R., Zhu, T., Parajuli, S., Guo, M., et~al.: The many faces of robustness: A
  critical analysis of out-of-distribution generalization. In: ICCV (2021)

\bibitem{hendrycks2018benchmarking}
Hendrycks, D., Dietterich, T.: Benchmarking neural network robustness to common
  corruptions and perturbations. In: ICLR (2018)

\bibitem{hendrycks2021natural}
Hendrycks, D., Zhao, K., Basart, S., Steinhardt, J., Song, D.: Natural
  adversarial examples. In: CVPR (2021)

\bibitem{heo2021rethinking}
Heo, B., Yun, S., Han, D., Chun, S., Choe, J., Oh, S.J.: Rethinking spatial
  dimensions of vision transformers. In: ICCV (2021)

\bibitem{herrmann2021pyramid}
Herrmann, C., Sargent, K., Jiang, L., Zabih, R., Chang, H., Liu, C., Krishnan,
  D., Sun, D.: Pyramid adversarial training improves vit performance. arXiv
  preprint arXiv:2111.15121  (2021)

\bibitem{hou2022vision}
Hou, Q., Jiang, Z., Yuan, L., Cheng, M.M., Yan, S., Feng, J.: Vision
  permutator: A permutable mlp-like architecture for visual recognition. IEEE
  Transactions on Pattern Analysis and Machine Intelligence  (2022)

\bibitem{huang2016deep}
Huang, G., Sun, Y., Liu, Z., Sedra, D., Weinberger, K.Q.: Deep networks with
  stochastic depth. In: ECCV (2016)

\bibitem{huang2017arbitrary}
Huang, X., Belongie, S.: Arbitrary style transfer in real-time with adaptive
  instance normalization. In: ICCV (2017)

\bibitem{jiang2021all}
Jiang, Z.H., Hou, Q., Yuan, L., Zhou, D., Shi, Y., Jin, X., Wang, A., Feng, J.:
  All tokens matter: Token labeling for training better vision transformers.
  In: NeurIPS (2021)

\bibitem{lin2014microsoft}
Lin, T.Y., Maire, M., Belongie, S., Hays, J., Perona, P., Ramanan, D.,
  Doll{\'a}r, P., Zitnick, C.L.: Microsoft coco: Common objects in context. In:
  European conference on computer vision. pp. 740--755. Springer (2014)

\bibitem{liu2022devil}
Liu, H., Jiang, X., Li, X., Guo, A., Jiang, D., Ren, B.: The devil is in the
  frequency: Geminated gestalt autoencoder for self-supervised visual
  pre-training. arXiv preprint arXiv:2204.08227  (2022)

\bibitem{liu2019roberta}
Liu, Y., Ott, M., Goyal, N., Du, J., Joshi, M., Chen, D., Levy, O., Lewis, M.,
  Zettlemoyer, L., Stoyanov, V.: Roberta: A robustly optimized bert pretraining
  approach. arXiv preprint arXiv:1907.11692  (2019)

\bibitem{liu2021swin}
Liu, Z., Lin, Y., Cao, Y., Hu, H., Wei, Y., Zhang, Z., Lin, S., Guo, B.: Swin
  transformer: Hierarchical vision transformer using shifted windows. In: CVPR
  (2021)

\bibitem{liu2019feature}
Liu, Z., Liu, Q., Liu, T., Xu, N., Lin, X., Wang, Y., Wen, W.: Feature
  distillation: Dnn-oriented jpeg compression against adversarial examples. In:
  CVPR (2019)

\bibitem{loshchilov2018decoupled}
Loshchilov, I., Hutter, F.: Decoupled weight decay regularization. In: ICLR
  (2018)

\bibitem{madry2018towards}
Madry, A., Makelov, A., Schmidt, L., Tsipras, D., Vladu, A.: Towards deep
  learning models resistant to adversarial attacks. In: ICLR (2018)

\bibitem{oppenheim1979phase}
Oppenheim, A., Lim, J., Kopec, G., Pohlig, S.: Phase in speech and pictures.
  In: ICASSP (1979)

\bibitem{oppenheim1981importance}
Oppenheim, A.V., Lim, J.S.: The importance of phase in signals. Proceedings of
  the IEEE  \textbf{69}(5),  529--541 (1981)

\bibitem{park2022how}
Park, N., Kim, S.: How do vision transformers work? In: ICLR (2022)

\bibitem{paszke2019pytorch}
Paszke, A., Gross, S., Massa, F., Lerer, A., Bradbury, J., Chanan, G., Killeen,
  T., Lin, Z., Gimelshein, N., Antiga, L., et~al.: Pytorch: an imperative
  style, high-performance deep learning library. In: NeurIPS (2019)

\bibitem{piotrowski1982demonstration}
Piotrowski, L.N., Campbell, F.W.: A demonstration of the visual importance and
  flexibility of spatial-frequency amplitude and phase. Perception
  \textbf{11}(3),  337--346 (1982)

\bibitem{qiu2021end2end}
Qiu, H., Gong, D., Li, Z., Liu, W., Tao, D.: End2end occluded face recognition
  by masking corrupted features. IEEE Transactions on Pattern Analysis and
  Machine Intelligence  (2021)

\bibitem{radosavovic2020designing}
Radosavovic, I., Kosaraju, R.P., Girshick, R., He, K., Doll{\'a}r, P.:
  Designing network design spaces. In: CVPR (2020)

\bibitem{raghunathan2020understanding}
Raghunathan, A., Xie, S.M., Yang, F., Duchi, J., Liang, P.: Understanding and
  mitigating the tradeoff between robustness and accuracy. In: ICML (2020)

\bibitem{recht2019imagenet}
Recht, B., Roelofs, R., Schmidt, L., Shankar, V.: Do imagenet classifiers
  generalize to imagenet? In: ICML (2019)

\bibitem{steiner2021train}
Steiner, A., Kolesnikov, A., Zhai, X., Wightman, R., Uszkoreit, J., Beyer, L.:
  How to train your vit? data, augmentation, and regularization in vision
  transformers. arXiv preprint arXiv:2106.10270  (2021)

\bibitem{strudel2021segmenter}
Strudel, R., Garcia, R., Laptev, I., Schmid, C.: Segmenter: Transformer for
  semantic segmentation. In: ICCV (2021)

\bibitem{sun2017revisiting}
Sun, C., Shrivastava, A., Singh, S., Gupta, A.: Revisiting unreasonable
  effectiveness of data in deep learning era. In: ICCV (2017)

\bibitem{sun2021rethinking}
Sun, Z., Cao, S., Yang, Y., Kitani, K.M.: Rethinking transformer-based set
  prediction for object detection. In: ICCV (2021)

\bibitem{sweldens1998lifting}
Sweldens, W.: The lifting scheme: A construction of second generation wavelets.
  SIAM journal on mathematical analysis  \textbf{29}(2),  511--546 (1998)

\bibitem{touvron2021training}
Touvron, H., Cord, M., Douze, M., Massa, F., Sablayrolles, A., J{\'e}gou, H.:
  Training data-efficient image transformers \& distillation through attention.
  In: ICML (2021)

\bibitem{vaswani2017attention}
Vaswani, A., Shazeer, N., Parmar, N., Uszkoreit, J., Jones, L., Gomez, A.N.,
  Kaiser, {\L}., Polosukhin, I.: Attention is all you need. In: NeurIPS (2017)

\bibitem{wang2018cosface}
Wang, H., Wang, Y., Zhou, Z., Ji, X., Gong, D., Zhou, J., Li, Z., Liu, W.:
  Cosface: Large margin cosine loss for deep face recognition. In: CVPR (2018)

\bibitem{wang2019learning}
Wang, H., Ge, S., Lipton, Z.C., Xing, E.P.: Learning robust global
  representations by penalizing local predictive power. In: NeurIPS (2019)

\bibitem{wang2020high}
Wang, H., Wu, X., Huang, Z., Xing, E.P.: High-frequency component helps explain
  the generalization of convolutional neural networks. In: CVPR (2020)

\bibitem{wang2021anti}
Wang, P., Zheng, W., Chen, T., Wang, Z.: Anti-oversmoothing in deep vision
  transformers via the fourier domain analysis: From theory to practice. In:
  ICLR (2022)

\bibitem{wang2021pyramid}
Wang, W., Xie, E., Li, X., Fan, D.P., Song, K., Liang, D., Lu, T., Luo, P.,
  Shao, L.: Pyramid vision transformer: A versatile backbone for dense
  prediction without convolutions. In: ICCV (2021)

\bibitem{wang2022crossformer}
Wang, W., Yao, L., Chen, L., Lin, B., Cai, D., He, X., Liu, W.: Crossformer: A
  versatile vision transformer hinging on cross-scale attention. In: ICLR
  (2022)

\bibitem{wang2021end}
Wang, Y., Xu, Z., Wang, X., Shen, C., Cheng, B., Shen, H., Xia, H.: End-to-end
  video instance segmentation with transformers. In: CVPR (2021)

\bibitem{wen2016discriminative}
Wen, Y., Zhang, K., Li, Z., Qiao, Y.: A discriminative deep feature learning
  approach for face recognition. In: ECCV (2016)

\bibitem{rw2019timm}
Wightman, R.: Pytorch image models.
  \url{https://github.com/rwightman/pytorch-image-models} (2019).
  \doi{10.5281/zenodo.4414861}

\bibitem{wightman2021resnet}
Wightman, R., Touvron, H., J{\'e}gou, H.: Resnet strikes back: An improved
  training procedure in timm. arXiv preprint arXiv:2110.00476  (2021)

\bibitem{xiao2018unified}
Xiao, T., Liu, Y., Zhou, B., Jiang, Y., Sun, J.: Unified perceptual parsing for
  scene understanding. In: Proceedings of the European Conference on Computer
  Vision (ECCV). pp. 418--434 (2018)

\bibitem{xie2020adversarial}
Xie, C., Tan, M., Gong, B., Wang, J., Yuille, A.L., Le, Q.V.: Adversarial
  examples improve image recognition. In: CVPR (2020)

\bibitem{xie2022masked}
Xie, J., Li, W., Zhan, X., Liu, Z., Ong, Y.S., Loy, C.C.: Masked frequency
  modeling for self-supervised visual pre-training. arXiv preprint
  arXiv:2206.07706  (2022)

\bibitem{xie2022simmim}
Xie, Z., Zhang, Z., Cao, Y., Lin, Y., Bao, J., Yao, Z., Dai, Q., Hu, H.:
  Simmim: A simple framework for masked image modeling. In: CVPR (2022)

\bibitem{xu2020learning}
Xu, K., Qin, M., Sun, F., Wang, Y., Chen, Y.K., Ren, F.: Learning in the
  frequency domain. In: CVPR (2020)

\bibitem{xu2021fourier}
Xu, Q., Zhang, R., Zhang, Y., Wang, Y., Tian, Q.: A fourier-based framework for
  domain generalization. In: CVPR (2021)

\bibitem{yang2020learning}
Yang, F., Yang, H., Fu, J., Lu, H., Guo, B.: Learning texture transformer
  network for image super-resolution. In: CVPR (2020)

\bibitem{yang2021larnet}
Yang, X., Jia, X., Gong, D., Yan, D.M., Li, Z., Liu, W.: Larnet: Lie algebra
  residual network for face recognition. In: International Conference on
  Machine Learning. pp. 11738--11750. PMLR (2021)

\bibitem{yin2019fourier}
Yin, D., Gontijo~Lopes, R., Shlens, J., Cubuk, E.D., Gilmer, J.: A fourier
  perspective on model robustness in computer vision. In: NeurIPS (2019)

\bibitem{yuan2021tokens}
Yuan, L., Chen, Y., Wang, T., Yu, W., Shi, Y., Jiang, Z.H., Tay, F.E., Feng,
  J., Yan, S.: Tokens-to-token vit: Training vision transformers from scratch
  on imagenet. In: ICCV (2021)

\bibitem{yuan2021volo}
Yuan, L., Hou, Q., Jiang, Z., Feng, J., Yan, S.: Volo: Vision outlooker for
  visual recognition. arXiv preprint arXiv:2106.13112  (2021)

\bibitem{yun2019cutmix}
Yun, S., Han, D., Oh, S.J., Chun, S., Choe, J., Yoo, Y.: Cutmix: Regularization
  strategy to train strong classifiers with localizable features. In: ICCV
  (2019)

\bibitem{zeng2020learning}
Zeng, Y., Fu, J., Chao, H.: Learning joint spatial-temporal transformations for
  video inpainting. In: ECCV (2020)

\bibitem{zhang2019theoretically}
Zhang, H., Yu, Y., Jiao, J., Xing, E., El~Ghaoui, L., Jordan, M.: Theoretically
  principled trade-off between robustness and accuracy. In: ICML (2019)

\bibitem{zhang2018mixup}
Zhang, H., Cisse, M., Dauphin, Y.N., Lopez-Paz, D.: mixup: Beyond empirical
  risk minimization. In: ICLR (2018)

\bibitem{zhou2019semantic}
Zhou, B., Zhao, H., Puig, X., Xiao, T., Fidler, S., Barriuso, A., Torralba, A.:
  Semantic understanding of scenes through the ade20k dataset. International
  Journal of Computer Vision  \textbf{127}(3),  302--321 (2019)

\bibitem{zhou2018end}
Zhou, L., Zhou, Y., Corso, J.J., Socher, R., Xiong, C.: End-to-end dense video
  captioning with masked transformer. In: CVPR (2018)

\bibitem{zhu2020freelb}
Zhu, C., Cheng, Y., Gan, Z., Sun, S., Goldstein, T., Liu, J.: Freelb: Enhanced
  adversarial training for natural language understanding. In: ICLR (2020)

\bibitem{zhu2020deformable}
Zhu, X., Su, W., Lu, L., Li, B., Wang, X., Dai, J.: Deformable detr: Deformable
  transformers for end-to-end object detection. In: ICLR (2020)

\end{thebibliography}

\appendix

\vspace{3em}
\begin{center}
    \begin{Large}
        \textbf{Appendix}
    \end{Large}
\end{center}

\section{More Related Works}
Deep neural networks (DNNs) have been showing state-of-the-art performances in various vision tasks  \cite{he2016deep,wen2016discriminative,wang2018cosface,deng2019mutual,devlin2019bert,alquraishi2019alphafold,qiu2021end2end,yang2021larnet}, however, previous works mainly investigated DNNs in the spatial domain, and also it is still an open problem to understand their mechanisms. Recently, many researchers take the Fourier transformation as a tool to improve DNNs' performance or analyze their behaviors \cite{wang2020high,xu2020learning,wang2021anti}.
One recent work \cite{wang2020high} investigates the CNN models from a frequency perspective. It reveals that CNN
can exploit the high-frequency image components that are
not perceivable to human, and demonstrates that it helps the generalization behaviors of CNN models. The method in \cite{xu2020learning} pre-processes the inputs in the frequency domain to better preserve image information and achieve improved accuracy in various vision tasks. Inspired by properties of the Fourier transformation \cite{oppenheim1979phase,oppenheim1981importance,piotrowski1982demonstration}, Xu $et$ $al$. \cite{xu2021fourier} proposed a novel Fourier-based data
augmentation strategy called amplitude mix to help domain generalization. Also, there are some works studying ViT models from the frequency domain. Park $et$ $al$. demonstrated that \cite{park2022how} multi-head self-attentions in ViT models are low-pass filters, but convolutional layers are high-pass filters, and proposed a novel architecture to combine these two operations. Theoretically, Wang $et$ $al$. explained that \cite{wang2021anti} cascading self-attention blocks in ViT models is equivalent to repeatedly applying a low-pass filter. Besides, most recent works \cite{liu2022devil,xie2022masked} solved the self-supervised visual pre-training through the lens of the frequency domain.

\section{Theoretical Analysis}
Based on definitions of the low-pass filtering $\mathcal{M}_l^{S}$ and high-pass filtering $\mathcal{M}_h^{S}$ in Section 3, following \cite{wang2021anti}, we present Theorem \ref{theorem1} for the attention map $\bm{A}\!\in \!\mathbb{R}^{n \! \times \! n}$ produced by the softmax function in ViT models. It is proven based on the property of the matrix $\bm{A}$ and the details can be found in \cite{wang2021anti}. Theorem \ref{theorem1} reveals that the self-attention mechanism diminishes the high-frequency information with depth increasing, which further confirms our hypothesis. 
\begin{theorem} 
 $\rm{lim}$$_{k \rightarrow \infty} \frac{||\mathcal{M}_h^{\!n-1}(\bm{A}^k\bm{v})||_{2}}{||\mathcal{M}_l^{\!1}(\bm{A}^k\bm{v})||_{2}} \!=\!0$ for any vector $\bm{v} \in \mathbb{R}^{n}$, where $\bm{A}^k$ denotes the product of $\bm{A}$ with itself $k$ times.
\label{theorem1}
\end{theorem} 

\section{Algorithm Pipeline}

\begin{algorithm}[H]
\caption{Training a ViT model with \texttt{HAT} for one epoch} 
{\bf Input:} 
Training set $X=\{(\bm{x},\bm{y})\}$, model weights $\bm{\theta}$, learning rate $\tau$, maximum perturbation strength $\epsilon$, number of PGD steps $K$, PGD step size $\eta$
\begin{algorithmic}[1]
\For{minibatch $B \subset X$}
\State $\bm{\delta}_0 \leftarrow \bm{0}$
\For{$t=1 ... K$}
\State {\fontsize{8pt}{\baselineskip}\selectfont $\triangleright$ Compute gradients of model weights and perturbations simultaneously}
\If{$t=1$} 
    \State $\nabla_{\bm{\theta}}, \nabla_{\bm{\delta}} \leftarrow  \nabla  L\big(\bm{\theta}, \bm{x} + \bm{\delta}_0, \bm{y}\big)$
\Else
    \State $\nabla_{\bm{\theta}}, \nabla_{\bm{\delta}} \leftarrow   \nabla \frac{1}{K\!-\!1} \left[\alpha L (\bm{\theta}, \bm{x}\!+\!\bm{\delta}_{t-1}, \bm{y})\! + \!\beta L_{kl}(\bm{\theta}, \bm{x}\!+\!\bm{\delta}_{t-1}, \bm{x})\right]$
\EndIf 

\State $\bm{g}_t \!\leftarrow \!\bm{g}_{t-1}\! +\! \nabla_{\bm{\theta}}$ 
~~~~~~~~~~~~~~~~~~~~~~~~~~~~~~{\fontsize{8pt}{\baselineskip}\selectfont $\triangleright$ Accumulate gradients of model weights}
\State $\bm{\delta}_t \!\leftarrow \! \text{clip}(\bm{\delta}_{t\!-\!1}\! + \eta \cdot \text{sign}(\nabla_{\!\bm{\delta}}), -\epsilon, \epsilon)$ 
~ ~~~~~~~~~~~~~~~~~~{\fontsize{8pt}{\baselineskip}\selectfont $\triangleright$ Update and clip perturbations}
\EndFor
\State $\bm{\theta} \leftarrow \bm{\theta}-\tau  \cdot \bm{g}_K$ ~~~~~~~~~~~~~~~~~~~~~~~~~~~~~~~~~~~~~~~~~~~~~~~~~~~~~~~{\fontsize{8pt}{\baselineskip}\selectfont $\triangleright$ Update model weights}
\EndFor
\end{algorithmic}
\label{alg:hat}
\end{algorithm}

\section{Implementation Details}

\subsection{Combining \texttt{HAT} with Knowledge Distillation}
We combine \texttt{HAT} with knowledge distillation by optimizing the loss function, which is obtained by replacing the cross-entropy loss in Eq. (4) with the distillation loss used in DeiT \cite{touvron2021training}, as follows:
\begin{equation*}
\centering
\mathbb{E}_{(\bm{x},\bm{y})\! \sim \!\mathcal{D}} \!\! \left [  L_{dist}\big(\bm{\theta}, \bm{x}, \bm{y}, \bm{y}_t\big) \!+\! \underset{||\bm{\delta}||_\infty\! \leqslant \epsilon}{\max} \Big( \alpha L_{dist} \big(\bm{\theta}, \bm{x}\!+\!\bm{\delta}, \bm{y}, \bm{y}_t\big)\! + \!\beta L_{kl}\big(\bm{\theta}, \bm{x}\!+\!\bm{\delta}, \bm{x}\big)\Big) \right ],
\label{eq:obj}
\end{equation*}
where $L_{dist}\big(\bm{\theta}, \bm{x}, \bm{y}, \bm{y}_t\big)\!=\!\frac{1}{2}\text{CE}(f_{\bm{\theta}}(\bm{x}), \bm{y})+\frac{1}{2}\text{CE}(f_{\bm{\theta}}(\bm{x}), \bm{y}_t)$ is the distillation loss and $\bm{y}_t$ is the hard decision of the teacher model. We also keep the optimization and all hyper-parameters unchanged.

\subsection{Object Detection and Instance Segmentation}
We take three variants of Swin Transformer trained without and with our \texttt{HAT} as pre-trained models to evaluate the performance in object detection and instance segmentation. The experiments are conducted on COCO 2017 \cite{lin2014microsoft}, which contains 118K training, 5K validation, and 20K test-dev images. We use the Cascade Mask R-CNN object detection framework \cite{cai2018cascade,he2017mask} with multi-scale training (resizing the input such that the shorter side is between 480 and 800 while the longer side is at most 1,333), AdamW optimizer (initial learning rate of 0.0001, weight decay of 0.05, and batch size of 16), and 3x schedule (36 epochs). Our implementation is based on Swin Transformer and more details can be found in \cite{liu2021swin}.

\subsection{Semantic Segmentation}
We also use Swin Transformer to evaluate the performance in semantic segmentation. We report results on the widely-used segmentation benchmark ADE20K~\cite{zhou2019semantic}, where ADE20K contains 25K images in total, including 20K images for training, 2K images for validation, and 3K images for test. And the UperNet \cite{xiao2018unified} is selected as the segmentation framework. In training, we follow the setup of the original paper \cite{liu2021swin}. Specifically, we utilize the AdamW optimizer with an initial learning rate of $6\times10^{-5}$ and a weight decay of 0.01, and we set the linear learning schedule with a minimum learning rate of $5\times10^{-6}$. Models are trained on 8 GPUs with 2 images per GPU for 160K iterations. In inference, we perform multi-scale test with interpolation rates of $[0.75, 1.0, 1.25, 1.5, 1.75]$.

\section{More Frequency Analysis}

\begin{figure}[t]
    \centering
    \subfigure[Low-pass Filtering]{
    \includegraphics[width=0.42\textwidth]{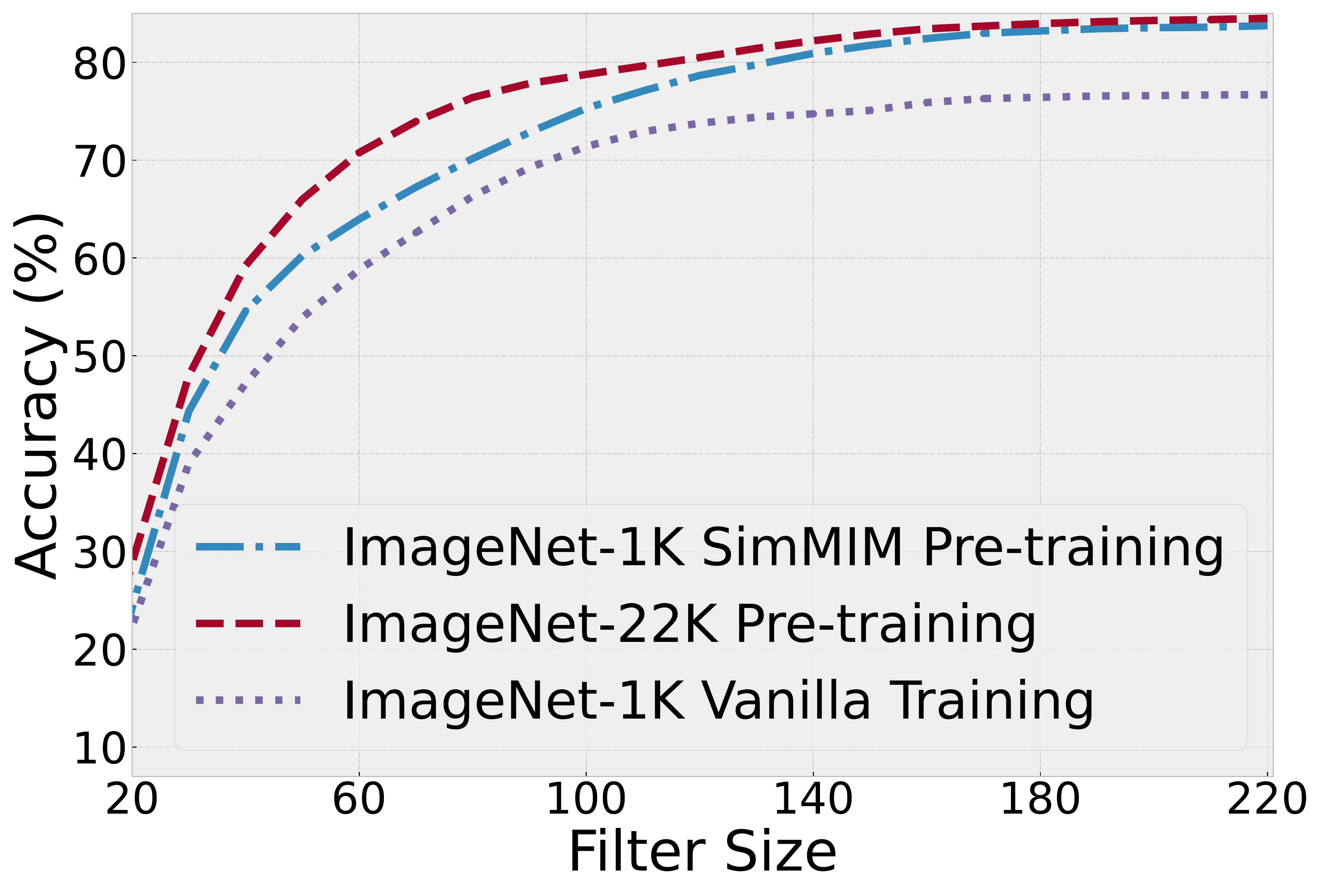}}
    \quad
    \subfigure[High-pass Filtering]{
    \includegraphics[width=0.42\textwidth]{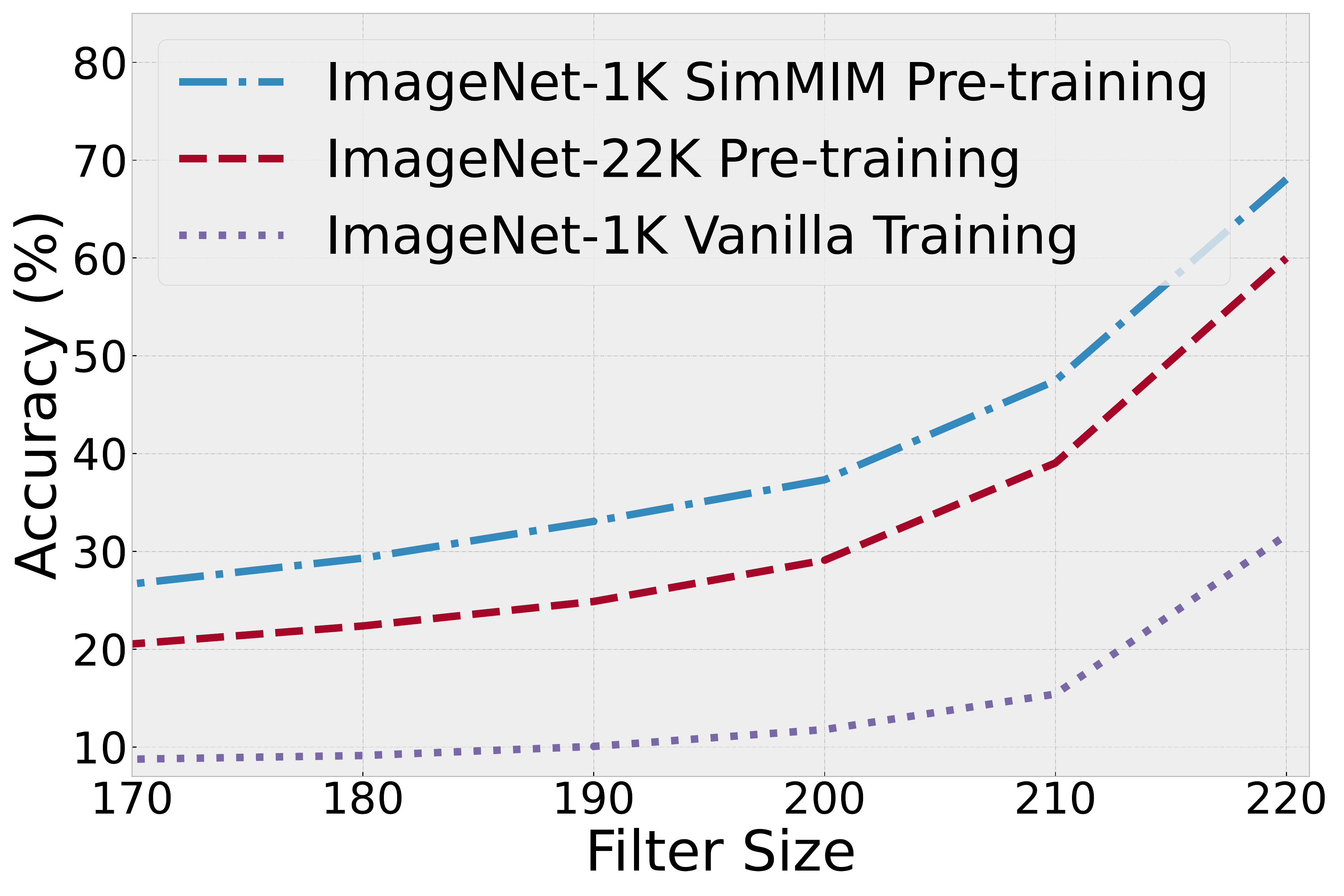}}
\setlength{\abovecaptionskip}{-0pt}
\setlength{\belowcaptionskip}{-20pt}
	\caption{Comparison of vanilla training and two pre-training methods on low- and high-pass filtered validation set with different filter sizes. The top-1 accuracy of ImageNet-1K Vanilla Training, ImageNet-22K Pre-training, and ImageNet-1K SimMIM Pre-training on the ImageNet validation set is 76.7\%, 84.5\%, and 83.8\%, respectively. }
	\label{fig:vit_pretrain}
\end{figure}

\subsection{Large-scale Pre-training}
 
We analyze supervised (i.e., ImageNet-22K) and self-supervised (i.e., SimMIM \cite{xie2022simmim}) pre-training in Figure \ref{fig:vit_pretrain}. It shows that both are helpful for exploiting high-frequency components, especially for SimMIM.

Besides, we directly use \texttt{HAT} in fine-tuning under the SimMIM pre-training setting. Specifically, we perform
adversarial training in the first 80 epochs for ViT-b and normal training in the rest 20 epochs, and keep other settings unchanged.
Training with \texttt{HAT} results in an 83.9\% top-1 accuracy, which is better than an 83.8\% top-1 accuracy reported in its paper and an 83.6\% top-1 accuracy in our reproduction without \texttt{HAT}.

\subsection{Knowledge Distillation}

\begin{figure}[t]
    \centering
    \subfigure[Low-pass Filtering]{
    \includegraphics[width=0.42\textwidth]{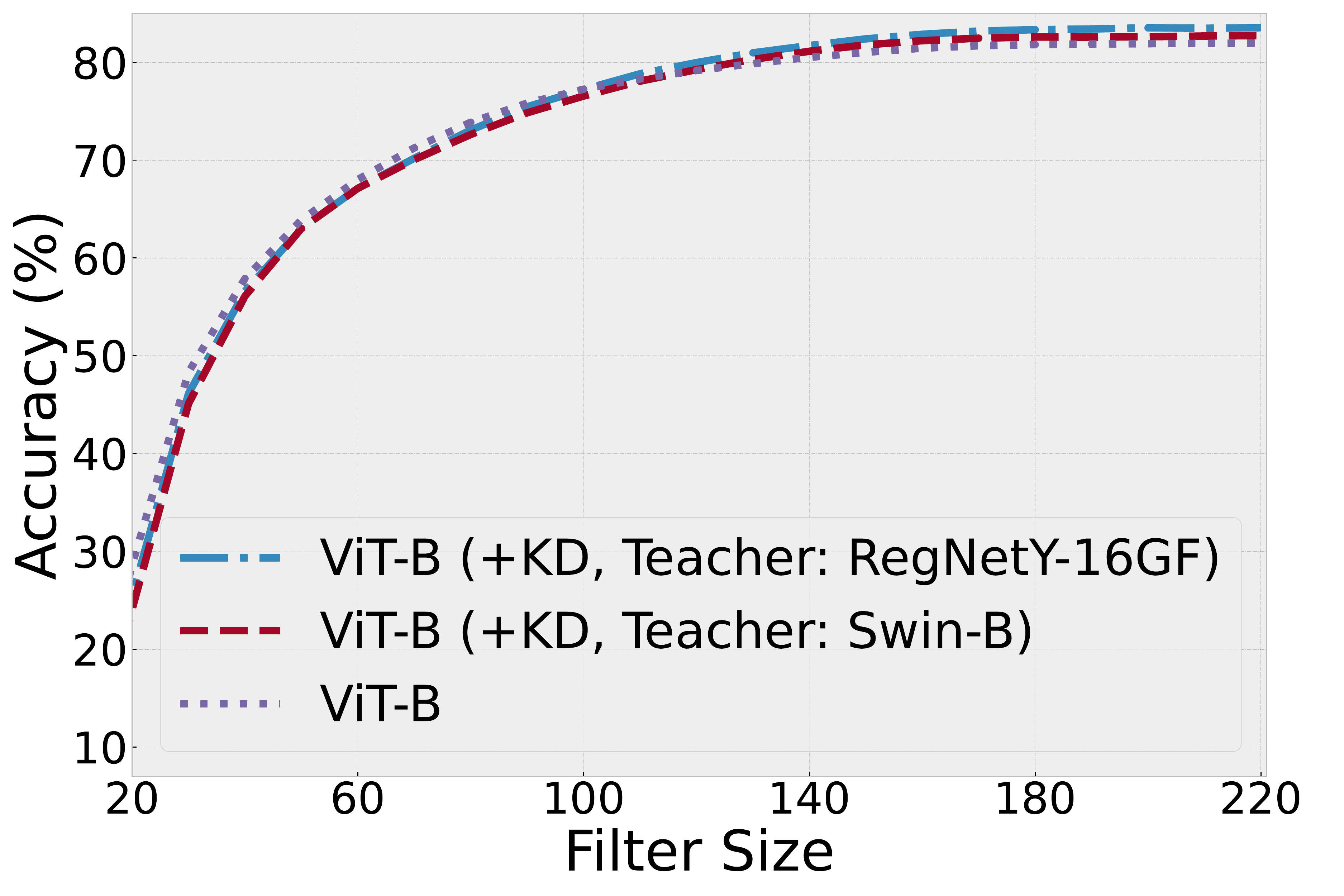}}
    \quad
    \subfigure[High-pass Filtering]{
    \includegraphics[width=0.42\textwidth]{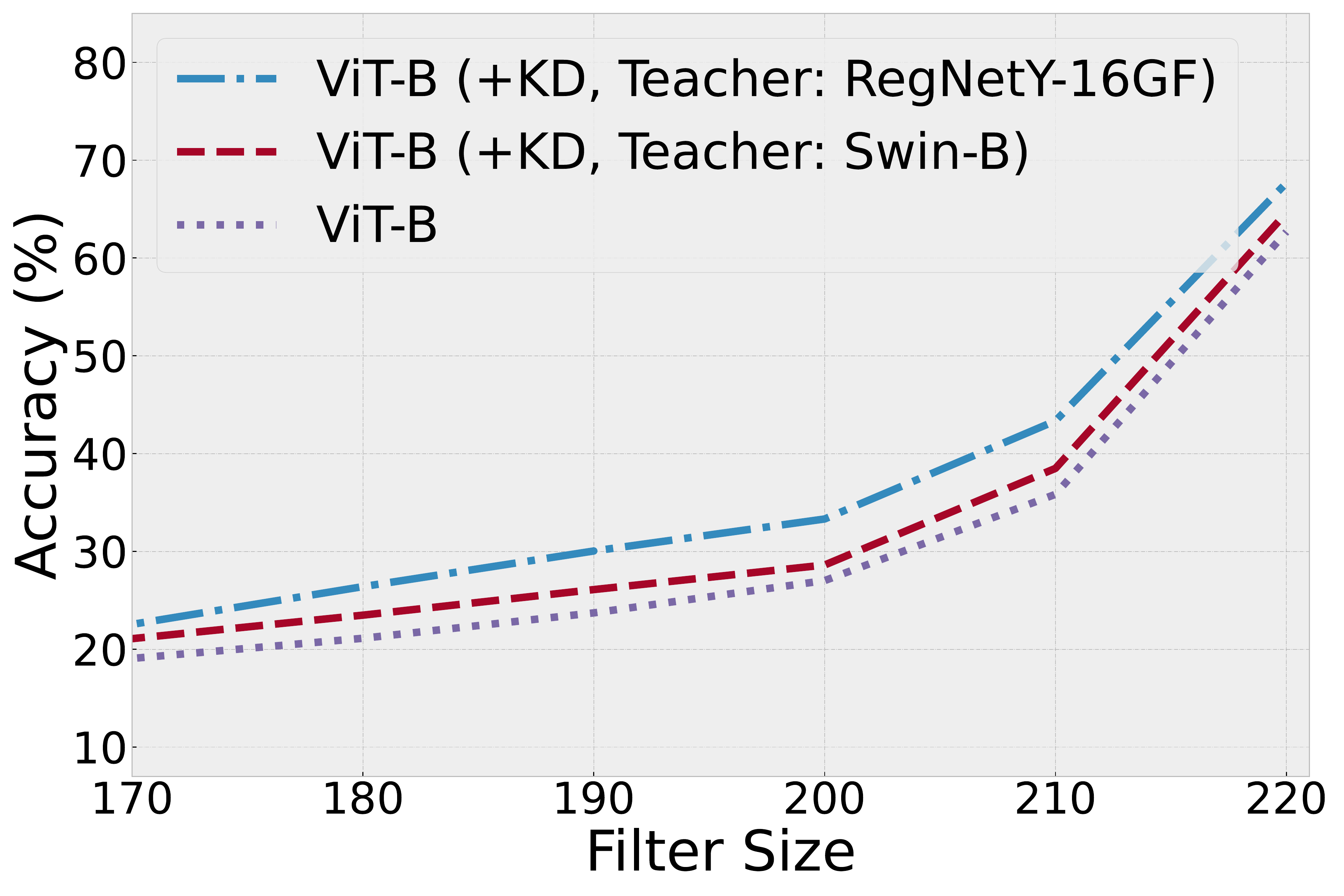}}
\setlength{\abovecaptionskip}{-0pt}
\setlength{\belowcaptionskip}{-20pt}
	\caption{Comparison of ViT-B and ViT-B (+KD) with different teacher models on low- and high-pass filtered validation set with different filter sizes. The top-1 accuracy of ViT-B, ViT-B (+KD, Teacher: Swin-B), and ViT-B (+KD, Teacher: RegNetY-16GF) on the ImageNet validation set is 82.0\%, 82.8\%, and 83.6\%, respectively.}
	\label{fig:vit_kd}
\end{figure}

Here we also analyze the knowledge distillation with CNN ($i.e.$, RegNetY-16GF) and ViT ($i.e.$, Swin-B) teacher in Figure \ref{fig:vit_kd}. It indicates that, compared to the ViT teacher, the CNN teacher is more helpful for ViT-B to capture the high-frequency components.

\end{document}